\documentclass[twoside,11pt,preprint]{article}

\usepackage{blindtext}

\usepackage{jmlr2e}

\usepackage{paralist,amsmath, amssymb}
\usepackage{algorithm}
\usepackage{algorithmic}
\usepackage{array}
\usepackage{makecell}
\usepackage{multirow}

\def \ze {\mathbf{0}}

\def \K {\mathcal{K}}
\def \L {\mathcal{L}}
\def \G {\mathcal{G}}

\def \C {\mathcal{C}}

\def \d {\mathbf{d}}
\def \E {\mathbb{E}}
\def \x {\mathbf{x}}
\def \y {\mathbf{y}}
\def \g {\mathbf{g}}
\def \z {\mathbf{z}}

\def \uu {\mathbf{u}}
\def \vv {\mathbf{v}}
\def \w {\mathbf{w}}

\DeclareMathOperator*{\ii}{in}
\DeclareMathOperator*{\oo}{out}

\DeclareMathOperator*{\argmin}{argmin}

\newtheorem{thm}{Theorem}
\newtheorem{lem}{Lemma}

\newtheorem{assum}{Assumption}

\newtheorem{cor}{Corollary}

\usepackage{times}

\usepackage{lastpage}
\jmlrheading{xx}{2024}{1-\pageref{LastPage}}{xx/xx}{xx/xx}{xx-xxxx}{Yuanyu Wan, Tong Wei, Bo Xue, Mingli Song, and Lijun Zhang}

\ShortHeadings{Optimal and Efficient Algorithms for Decentralized Online Convex Optimization}{Wan, Wei, Xue, Song, and Zhang}
\firstpageno{1}

\begin{document}

\title{Optimal and Efficient Algorithms for Decentralized \\Online Convex Optimization}

\author{%
 \name Yuanyu Wan \email wanyy@zju.edu.cn\\
 \addr School of Software Technology, Zhejiang University, Ningbo, China\\
 \addr State Key Laboratory of Blockchain and Data Security, Zhejiang University, Hangzhou, China
 \AND
 \name Tong Wei \email weit@seu.edu.cn\\
 \addr School of Computer Science and Engineering, Southeast University, Nanjing, China
 \AND
 \name Bo Xue \email boxue4-c@my.cityu.edu.hk\\
 \addr Department of Computer Science, City University of Hong Kong, Hong Kong, China
 \AND
 \name Mingli Song \email brooksong@zju.edu.cn\\
 \addr State Key Laboratory of Blockchain and Data Security, Zhejiang University, Hangzhou, China
 \AND
  \name Lijun Zhang \email zhanglj@lamda.nju.edu.cn\\
 \addr National Key Laboratory for Novel Software Technology, Nanjing University, Nanjing, China\\
 \addr School of Artificial Intelligence, Nanjing University, Nanjing, China%
}

\editor{}

\maketitle

\begin{abstract}%
We investigate decentralized online convex optimization (D-OCO), in which a set of local learners are required to minimize a sequence of global loss functions using only local computations and communications. Previous studies have established $O(n^{5/4}\rho^{-1/2}\sqrt{T})$ and ${O}(n^{3/2}\rho^{-1}\log T)$ regret bounds for convex and strongly convex functions respectively, where $n$ is the number of local learners, $\rho<1$ is the spectral gap of the communication matrix, and $T$ is the time horizon. However, there exist large gaps from the existing lower bounds, i.e., $\Omega(n\sqrt{T})$ for convex functions and $\Omega(n)$ for strongly convex functions. To fill these gaps, in this paper, we first develop a novel D-OCO algorithm that can respectively reduce the regret bounds for convex and strongly convex functions to $\tilde{O}(n\rho^{-1/4}\sqrt{T})$ and $\tilde{O}(n\rho^{-1/2}\log T)$. The primary technique is to design an online accelerated gossip strategy that enjoys a faster average consensus among local learners. Furthermore, by carefully exploiting spectral properties of a specific network topology, we enhance the lower bounds for convex and strongly convex functions to $\Omega(n\rho^{-1/4}\sqrt{T})$ and $\Omega(n\rho^{-1/2}\log T)$, respectively. These results suggest that the regret of our algorithm is nearly optimal in terms of $T$, $n$, and $\rho$ for both convex and strongly convex functions. Finally, we propose a projection-free variant of our algorithm to efficiently handle practical applications with complex constraints. Our analysis reveals that the projection-free variant can achieve ${O}(nT^{3/4})$ and ${O}(nT^{2/3}(\log T)^{1/3})$ regret bounds for convex and strongly convex functions with nearly optimal $\tilde{O}(\rho^{-1/2}\sqrt{T})$ and $\tilde{O}(\rho^{-1/2}T^{1/3}(\log T)^{2/3})$ communication rounds, respectively. 
\end{abstract}


\section{Introduction}
Decentralized online convex optimization (D-OCO) \citep{DAOL_TKDE,D-ODA,wenpeng17,Wan-ICML-2020,Wan-22-JMLR} is a powerful learning framework for distributed applications with streaming data, such as distributed tracking in sensor networks \citep{Distrbuted02,DSN_book} and online packet routing \citep{Awerbuch04,Awerbuch2008}. Specifically, it can be formulated as a repeated game between an adversary and a set of local learners numbered by $1,\dots, n$ and connected by a network, where the network is defined by an undirected graph $\mathcal{G}=([n],E)$ with the edge set $E\subseteq [n]\times [n]$. In the $t$-th round, each learner $i\in [n]$ first chooses a decision $\x_i(t)$ from a convex set $\K\subseteq \mathbb{R}^d$, and then receives a convex loss function $f_{t,i}(\x):\K\to\mathbb{R}$ selected by the adversary. The goal of each learner $i$ is to minimize the regret in terms of the global function $f_t(\x)=\sum_{j=1}^nf_{t,j}(\x)$ at each round $t$, i.e.,
\begin{equation}
\label{reg_DOCO}
R_{T,i}=\sum_{t=1}^Tf_{t}(\x_i(t))-\min\limits_{\x\in\K}\sum_{t=1}^Tf_{t}(\x)
\end{equation}
where $T$ denotes the time horizon.

Note that in the special case with $n=1$, D-OCO reduces to the classical online convex optimization (OCO) \citep{Online:suvery,Hazan2016}. There already exist many online algorithms with optimal regret bounds for convex and strongly convex functions, e.g., online gradient descent (OGD) \citep{Zinkevich2003}. However, these algorithms cannot be applied to the general D-OCO problem, because they need direct access to the global function $f_t(\x)$, which is unavailable for the local learners. To be precise, there exist communication constraints in D-OCO: the learner $i$ only has local access to the function $f_{t,i}(\x)$, and can only communicate with its immediate neighbors via a single step of the gossip protocol \citep{Xiao-Gossip04,Gossip06} based on a weight matrix $P\in\mathbb{R}^{n\times n}$ at each round.\footnote{More specifically, the essence of a single gossip step is to compute a weighted average of some parameters of these local learners based on the matrix $P$. Moreover, following previous studies \citep{DAOL_TKDE,D-ODA}, $P$ is given beforehand, instead of being a choice of the algorithm.} To address this limitation, the pioneering work of \citet{DAOL_TKDE} extends OGD into the D-OCO setting, and achieves $O(n^{5/4}\rho^{-1/2}\sqrt{T})$ and ${O}(n^{3/2}\rho^{-1}\log T)$ regret bounds for convex and strongly convex functions respectively, where $\rho<1$ is the spectral gap of $P$. The key idea is to first apply a standard gossip step \citep{Xiao-Gossip04} over the decisions of these local learners, and then perform a gradient descent step based on the local function. Later, there has been a growing research interest in developing and analyzing D-OCO algorithms based on the standard gossip step for various scenarios \citep{D-ODA,wenpeng17,NIPS20-random-DOCO,Wan-ICML-2020,Wan2021arXiv,Wan-22-JMLR,Wang-AAAI-2023}. However, the best regret bounds for D-OCO with convex and strongly convex functions remain unchanged. Moreover, there exist large gaps from the lower bounds recently established by \citet{Wan-22-JMLR}, i.e., $\Omega(n\sqrt{T})$ for convex functions and $\Omega(n)$ for strongly convex functions.

To fill these gaps, this paper first proposes a novel D-OCO algorithm that can achieve a regret bound of $\tilde{O}(n\rho^{-1/4}\sqrt{T})$ for convex functions and an improved regret bound of $\tilde{O}(n\rho^{-1/2}\log T)$ for strongly convex functions, respectively.\footnote{We use the $\tilde{O}(\cdot)$ notation to hide constant factors as well as polylogarithmic factors in $n$, but not in $T$.}
Different from previous D-OCO algorithms that rely on the standard gossip step, we make use of an accelerated gossip strategy \citep{Acc_Gossip} to weaken the impact
of decentralization on the regret. In the studies of offline and stochastic optimization, it is well-known that the accelerated strategy enjoys a faster average consensus among decentralized nodes \citep{Lu-ICML-21,YeAccGossip23,Ye2020}. However, applying the accelerated strategy to D-OCO is more challenging because it requires multiple communications in each round, which violates the communication protocol of D-OCO. To tackle this issue, we design an online accelerated gossip strategy by further incorporating a blocking update mechanism, which allows us to allocate the communications required by each update into every round of a block. Furthermore, we establish nearly matching lower bounds of $\Omega(n\rho^{-1/4}\sqrt{T})$ and $\Omega(n\rho^{-1/2}\log T)$ for convex and strongly convex functions, respectively. Compared with existing lower bounds \citep{Wan-22-JMLR}, our bounds additionally uncover the effect of the spectral gap, by carefully exploiting spectral properties of a specific network topology. Table \ref{tab1} provides a comparison of our results on the optimality of D-OCO with those of previous studies.
\begin{table}[t]
 \centering
 \caption{Summary of our results and the best previous results on the optimality of D-OCO. Abbreviations: convex $\to$ cvx, strongly convex $\to$ scvx.}
 \label{tab1}
 \begin{tabular}{|c|c|c|c|c|}
    \hline
    $f_{t,i}(\cdot)$ & Source & Upper Bound & Lower Bound &Regret Gap\\
    \hline
    \multirow{2}{*}{cvx} &Previous studies & \makecell{$O(n^{5/4}\rho^{-1/2}\sqrt{T})$\\\citet{DAOL_TKDE};\\\citet{D-ODA}} & \makecell{$\Omega(n\sqrt{T})$\\\citet{Wan-22-JMLR}} & $O(n^{1/4}\rho^{-1/2})$ \\
    \cline{2-5}
    ~ & This work & \makecell{${O}(n\rho^{-1/4}\sqrt{T\log n})$\\Corollary \ref{corollary-convex_upper}} & \makecell{$\Omega(n\rho^{-1/4}\sqrt{T})$\\Theorem \ref{thm_lowerB}} & ${O}(\sqrt{\log n})$ \\
    \hline
    \multirow{2}{*}{scvx} &Previous studies & \makecell{${O}(n^{3/2}\rho^{-1}\log T)$\\\citet{DAOL_TKDE}} & \makecell{$\Omega(n)$\\\citet{Wan-22-JMLR}} & ${O}(n^{1/2}\rho^{-1}\log T)$ \\
    \cline{2-5}
    ~&  This work & \makecell{${O}(n\rho^{-1/2}(\log n)\log T)$\\Corollary \ref{corollary-sconvex_upper}}& \makecell{ $\Omega(n\rho^{-1/2}\log T)$ \\Theorem \ref{imp-thm_lowerB-sc}}
    & ${O}(\log n)$ \\
    \hline
 \end{tabular}
\end{table}
\begin{table}[t]
 \centering
 \caption{Summary of our projection-free algorithm and the best existing projection-free algorithm for D-OCO. Abbreviations: convex $\to$ cvx, strongly convex $\to$ scvx.}
 \label{tab2}
 \begin{tabular}{|c|c|c|c|}
    \hline
    $f_{t,i}(\cdot)$ & Source & Regret & Communication Rounds\\
    \hline
    \multirow{3}{*}{cvx} &\citet{Wan-22-JMLR} & {${O}(n^{5/4}\rho^{-1/2}T^{3/4})$ } & {$O(\sqrt{T})$} \\
    \cline{2-4}
    ~ & \multirow{2}{*}{Corollary \ref{corollary-convex_upper-projection-free}} & {${O}(n\rho^{-1/4}T^{3/4}\sqrt{\log n})$} & {$O(\sqrt{T})$} \\
     \cline{3-4}
    ~ & ~& {${O}(nT^{3/4})$} & {$O(\rho^{-1/2}\sqrt{T}\log n)$} \\
    \hline
    \multirow{3}{*}{scvx} &\citet{Wan-22-JMLR} & ${O}(n^{3/2}\rho^{-1}T^{2/3}(\log T)^{1/3})$ & ${O}(T^{1/3}(\log T)^{2/3})$ \\
    \cline{2-4}
    ~& \multirow{2}{*}{Corollary \ref{corollary-sconvex_upper-projection-free}} & ${O}(n\rho^{-1/2}T^{2/3}(\log T)^{1/3}\log n)$& ${O}(T^{1/3}(\log T)^{2/3})$\\
    \cline{3-4}
    ~ & ~& ${O}(nT^{2/3}(\log T)^{1/3})$ & ${O}(\rho^{-1/2}T^{1/3}(\log T)^{2/3}\log n)$\\
    \hline
 \end{tabular}
\end{table}

Finally, we notice that each learner of our algorithm needs to perform a projection operation per round to ensure the feasibility of its decision, which could be computationally expensive in applications with complex constraints. For example, in online collaborative filtering \citep{Hazan2012}, the decision set consists of all matrices with bounded trace norm, and the corresponding projection needs to compute singular value decomposition of a matrix. To tackle this computational bottleneck, we propose a projection-free variant of our algorithm by replacing the projection operation with a more efficient linear optimization step. Analysis reveals that our projection-free algorithm can achieve a regret bound of ${O}(nT^{3/4})$ with only $\tilde{O}(\rho^{-1/2}\sqrt{T})$ communication rounds for convex functions, and a better regret bound of ${O}(nT^{2/3}(\log T)^{1/3})$  with fewer $\tilde{O}(\rho^{-1/2}T^{1/3}(\log T)^{2/3})$ communication rounds for strongly convex functions, respectively. In contrast, the state-of-the-art projection-free D-OCO algorithm \citep{Wan-22-JMLR} only achieves worse ${O}(n^{5/4}\rho^{-1/2}T^{3/4})$ and ${O}(n^{3/2}\rho^{-1}T^{2/3}(\log T)^{1/3})$ regret bounds for convex and strongly convex functions respectively, though the number of required communication rounds is less, i.e., ${O}(\sqrt{T})$ for convex functions and ${O}(T^{1/3}(\log T)^{2/3})$ for strongly convex functions. Moreover, even if using the same number of communication rounds as \citet{Wan-22-JMLR}, we show that the regret bounds of our projection-free algorithm only degenerate to $\tilde{O}(n\rho^{1/4}T^{3/4})$ and $\tilde{O}(n\rho^{1/2}T^{2/3}(\log T)^{1/3})$, which are still tighter than those of \citet{Wan-22-JMLR}, respectively. It is worth noting that although these regret bounds of our projection-free algorithm no longer match the aforementioned lower bounds, we further extend the latter to demonstrate that the number of required communication rounds is nearly optimal for achieving these regret bounds. Table \ref{tab2} provides a comparison of our projection-free algorithm and that of \citet{Wan-22-JMLR}.

A preliminary version of this paper was presented at the 37th Annual Conference on Learning Theory in 2024 \citep{COLT-24-wan}. In this paper, we have significantly enriched the preliminary version by adding the following extensions.
\begin{compactitem}
\item Different from \citet{COLT-24-wan} that propose two algorithms to deal with convex and strongly convex functions respectively, we unify them into a single one, and provide a unified analysis to recover their regret bounds.
\item For D-OCO with strongly convex functions, we improve the $\Omega(n\rho^{-1/2})$ lower bound in \citet{COLT-24-wan} to $\Omega(n\rho^{-1/2}\log T)$, which now can recover the classical $\Omega(\log T)$ lower bound for OCO with strongly convex functions \citep{Abernethy08,Epoch-GD}.
\item We propose a projection-free variant of our algorithm to efficiently handle complex constraints. It improves the regret of the state-of-the-art projection-free D-OCO algorithm \citep{Wan-22-JMLR} while maintaining the same number of communication rounds.
\item We establish $\Omega(n\rho^{-1/4}T/\sqrt{C})$ and $\Omega(n\rho^{-1/2}T/C)$ lower bounds for convex and strongly convex functions in a more challenging setting with only $C$ communication rounds, which imply that the communication complexity of our projection-free variant is nearly optimal.
\item We demonstrate that the analysis of some existing algorithms \citep{D-ODA,Wan-22-JMLR} can be refined to reduce the dependence of their current regret bounds on $n$ to only $\tilde{O}(n)$. Although the refined regret is still not optimal, it may be of independent interest.
\end{compactitem}

\section{Related Work}
In this section, we briefly review the related work on D-OCO, including the special case with $n=1$ and the general case.

\subsection{Special D-OCO with $n=1$}
D-OCO with $n=1$ reduces to the classical OCO problem, which dates back to the seminal work of \citet{Zinkevich2003}. Over the past decades, this problem has been extensively studied, and various
algorithms with optimal regret have been presented for convex and strongly convex functions, respectively \citep{Zinkevich2003,Shai07,Hazan_2007,Abernethy08,Epoch-GD}. The closest one to this paper is follow-the-regularized-leader (FTRL) \citep{Shai07}, which updates the decision (omitting the subscript of the learner 1 for brevity) as
\begin{equation}
\label{FTRL}
\begin{split}
\x(t+1) = \argmin_{\x\in\K}\sum_{i=1}^t\left\langle\nabla f_i(\x(i)),\x\right\rangle+\frac{1}{\eta}\|\x\|_2^2
\end{split}
\end{equation}
where $\eta$ is a parameter. By tuning $\eta$ appropriately, it can achieve an optimal $O(\sqrt{T})$ regret bound for convex functions. Note that this algorithm is also known as dual averaging, especially in the filed of offline and stochastic optimization \citep{Dual-Averaging09,NIPS2009_3882}. Moreover, \citet{Hazan_2007} have proposed a variant of \eqref{FTRL} for $\alpha$-strongly convex functions, which makes the following update
\begin{equation}
\label{FTAL}
\begin{split}
\x(t+1) =&\argmin_{\x\in\K}\sum_{i=1}^t\left(\left\langle\nabla f_i(\x(i)),\x\right\rangle+\frac{\alpha}{2}\|\x-\x(i)\|_2^2\right)\\
=&\argmin_{\x\in\K}\sum_{i=1}^t\langle\nabla f_i(\x(i))-\alpha\x(i), \x\rangle+\frac{t\alpha}{2}\|\x\|_2^2.
\end{split}
\end{equation}
It is named as follow-the-approximate-leader (FTAL), and can achieve an optimal $O(\log T)$ regret bound for strongly convex functions. 

Nonetheless, as first noticed by \citet{Hazan2012}, these optimal algorithms explicitly or implicitly require a projection operation per round, which could become a computational bottleneck for complex decision sets. To address this issue, \citet{Hazan2012} propose a projection-free variant of FTRL, and establish an $O(T^{3/4})$ regret bound for convex functions. Their key idea is to approximately solve \eqref{FTRL} with only one linear optimization step, which could be much more efficiently carried out than the projection operation.  Due to the efficiency, there has been a growing research interest in developing projection-free OCO algorithms \citep{Garber16,kevy_smooth,Hazan20,Garber_SOFW,SC_OFW,DY_OFW,Garber-COLT22,Garber-COLT23,COLT-23-wan}. The most related work to this paper is \citet{SC_OFW}, in which a projection-free variant of FTAL is proposed to achieve an improved regret bound of  $O(T^{2/3})$ for strongly convex functions.

\subsection{General D-OCO with $n\geq 2$}
\label{sec2-2}
D-OCO is a generalization of OCO with $n\geq 2$ local learners in the network defined by an undirected graph $\mathcal{G}=([n],E)$. The main challenge of D-OCO is that each learner $i\in [n]$ is required to minimize the regret in terms of the global function $f_t(\x)=\sum_{j=1}^nf_{t,j}(\x)$, i.e., $R_{T,i}$ in \eqref{reg_DOCO}, but except the direct access to $f_{t,i}(\x)$, it can only estimate the global information from the gossip communication occurring via the weight matrix $P$. To tackle this challenge, \citet{DAOL_TKDE}  propose a decentralized variant of OGD (D-OGD) by first applying the standard gossip step \citep{Xiao-Gossip04} over the decisions of these local learners, and then performing a gradient descent step according to the local function. For convex and strongly convex functions, D-OGD can achieve $O(n^{5/4}\rho^{-1/2}\sqrt{T})$ and ${O}(n^{3/2}\rho^{-1}\log T)$ regret bounds, respectively. Later, \citet{D-ODA} propose a decentralized variant of FTRL (D-FTRL), which performs the following update
\begin{equation}
 \label{DFTRL}
\begin{split}
&\z_{i}(t+1)=\sum_{j\in N_i}P_{ij}\z_j(t)+\nabla f_{t,i}(\x_i(t))\\
&\x_i(t+1) = \argmin_{\x\in\K}\left\langle\z_{i}(t+1),\x\right\rangle+\frac{1}{\eta}\|\x\|_2^2
\end{split}
\end{equation}
for each learner $i$, where $N_i=\{j\in [n]|(i,j)\in E\}\cup\{i\}$ denotes the set including the immediate neighbors of the learner $i$ and itself. Notice that the cumulative gradients $\sum_{i=1}^t\nabla f_i(\x(i))$ utilized in \eqref{FTRL} is replaced by a local variable $\z_{i}(t+1)$ that is computed by first applying the standard gossip step over $\z_i(t)$ of these local learners and then adding the local gradient $\nabla f_{t,i}(\x_i(t))$. For convex functions, D-FTRL can also achieve the $O(n^{5/4}\rho^{-1/2}\sqrt{T})$ regret bound.

The first projection-free algorithm for D-OCO is proposed by \citet{wenpeng17}, and can be viewed as a combination of D-FTRL and linear optimization steps. For convex functions, analogous to the projection-free variant of FTRL \citep{Hazan2012} in OCO, this algorithm can achieve an $O(n^{5/4}\rho^{-1/2}T^{3/4})$ regret bound. After that, several improvements have been made to projection-free D-OCO \citep{Wan-ICML-2020,Wan2021arXiv,Wan-22-JMLR,Wang-AAAI-2023}. First, \citet{Wan-ICML-2020} demonstrate that by combining the projection-free algorithm in \citet{wenpeng17} with a blocking update mechanism, the number of communication rounds can be reduced from $O(T)$ to $O(\sqrt{T})$ while achieving the same $O(n^{5/4}\rho^{-1/2}T^{3/4})$ regret bound for convex functions. If the functions are strongly convex, \citet{Wan2021arXiv} propose a projection-free and decentralized variant of FTAL \citep{Hazan_2007}, which can enjoy an improved regret bound of ${O}(n^{3/2}\rho^{-1}T^{2/3}(\log T)^{1/3})$ with even fewer $O(T^{1/3}(\log T)^{2/3})$ communication rounds. Then, \citet{Wan-22-JMLR} unify these two algorithms into a single one that inherits the theoretical guarantees for both convex and strongly convex functions. Moreover, they also provide $\Omega(nT/\sqrt{C})$ and $\Omega(nT/C)$ lower regret bounds for any D-OCO algorithm with $C$ communication rounds by respectively considering convex and strongly convex functions. These lower bounds imply that the number of communication rounds required by their algorithm is (nearly) optimal in terms of $T$ for achieving their current regret bounds. Very recently, \citet{Wang-AAAI-2023} develop a randomized projection-free algorithm for D-OCO with smooth functions, and achieve an expected regret bound of $O(n^{5/4}\rho^{-1/2}T^{2/3})$ with $O(T^{2/3})$ communication rounds.

Additionally, we notice that if the projection operation is allowed, the unified algorithm in \citet{Wan-22-JMLR} can be simplified as performing the following update
\begin{equation}
 \label{DFTAL}
\begin{split}
&\z_{i}(t+1)=\sum_{j\in N_i}P_{ij}\z_j(t)+(\nabla f_{t,i}(\x_i(t))-\alpha\x_i(t))\\
&\x_i(t+1) = \argmin_{\x\in\K}\left\langle\z_{i}(t+1),\x\right\rangle+\frac{t\alpha}{2}\|\x\|_2^2+h\|\x\|_2^2
\end{split}
\end{equation}
for each learner $i$, where $\alpha$ and $h$ are two parameters. By setting $\alpha=0$ and $h=1/\eta$, \eqref{DFTAL} reduces to D-FTRL and thus can also enjoy the $O(n^{5/4}\rho^{-1/2}\sqrt{T})$ regret bound for convex functions. For $\alpha$-strongly convex functions, by simply setting $h=0$, \eqref{DFTAL} becomes a decentralized variant of FTAL because the local variable $\z_{i}(t+1)$ now is utilized to replace the cumulative information $\sum_{i=1}^t(\nabla f_i(\x(i))-\alpha\x(i))$ in \eqref{FTAL}. This subtle difference allows the algorithm to recover the ${O}(n^{3/2}\rho^{-1}\log T)$ regret bound for strongly convex functions (see Appendix \ref{appendix_last} for details). Therefore, \eqref{DFTAL} can be referred to as decentralized follow-the-generalized-leader (D-FTGL). Moreover, by setting $C=O(T)$, the communication-dependent lower bounds in \citet{Wan-22-JMLR} reduce to $\Omega(n\sqrt{T})$ and $\Omega(n)$ lower bounds for general D-OCO with convex and strongly convex functions, respectively. These results imply that the existing $O(n^{5/4}\rho^{-1/2}\sqrt{T})$ and ${O}(n^{3/2}\rho^{-1}\log T)$ upper bounds are (nearly) tight in terms of $T$. However, there still exist gaps in terms of $n$ and $\rho$ between these upper and lower bounds. Note that the value of $\rho$ reflects the connectivity of the network: a larger $\rho$ implies better connectivity, and could even be $\Omega(n^{-2})$ for ``poorly connected'' networks such as the $1$-connected cycle graph \citep{DADO2011}. Therefore, these gaps on $n$ and $\rho$ cannot be ignored, especially for large-scale distributed systems. In this paper, we fill these gaps up to polylogarithmic factors in $n$.

\paragraph{Discussions} Different from D-OCO, previous studies have proposed optimal algorithms for many different scenarios of decentralized offline and stochastic optimization \citep{Scaman-ICML17,Scaman-NIPS18,Scaman-JMLR19,Gorbunov-Arxiv-2020,Kovalev-NIPS20,Dvinskikh-Arxiv-21,Lu-ICML-21,SongAccGossip23,YeAccGossip23,Ye2020}. The closest work to this paper is \citet{Scaman-JMLR19}, which investigates decentralized offline optimization with convex and strongly convex functions. Let $\hat{\rho}$ be the normalized eigengap of $P$, which could be close to $\rho$. \citet{Scaman-JMLR19} have established optimal convergence rates of $O(\epsilon^{-2}+\epsilon^{-1}\hat{\rho}^{-1/2})$ and $O(\epsilon^{-1}+\epsilon^{-1/2}\hat{\rho}^{-1/2})$ to reach an $\epsilon$ precision for convex and strongly convex functions, respectively. However, it is worth noting that D-OCO is more challenging than the offline setting due to the change of functions. Actually, it is easy to apply a standard online-to-batch conversion \citep{O2B} of any D-OCO algorithm with regret $R_{T,i}$ to achieve an approximation error of $O(R_{T,i}/(nT))$ for decentralized offline optimization, but not vice versa. Moreover, one may notice that due to the online-to-batch conversion, it is possible to utilize existing lower bounds in the offline setting \citep{Scaman-JMLR19} to derive $\Omega(n\sqrt{T}+n\hat{\rho}^{-1/2})$ and $\Omega(n+n\hat{\rho}^{-1}T^{-1})$ lower bounds for the regret of D-OCO with convex and strongly convex functions, respectively. However, for D-OCO, it is common to consider the case where $T$ is much larger than other problem constants, and these lower bounds will reduce to the $\Omega(n\sqrt{T})$ and $\Omega(n)$ lower bounds specially established for D-OCO \citep{Wan-22-JMLR}. In addition, we want to emphasize that although the accelerated gossip strategy \citep{Acc_Gossip} has been widely utilized in these previous studies on decentralized offline and stochastic optimization, this paper is the first work to apply it in D-OCO.
\section{Preliminaries}
In this section, we introduce the necessary preliminaries including common assumptions and an algorithmic ingredient. Specifically, similar to previous studies on D-OCO \citep{DAOL_TKDE,D-ODA}, we introduce the following assumptions.
\begin{assum}
\label{assum5}
The communication matrix $P\in \mathbb{R}^{n\times n}$ is supported on the graph $\mathcal{G}=([n],E)$, symmetric, and doubly stochastic, which satisfies
\begin{compactitem}
\item $P_{ij}>0$ only if $(i,j)\in E$ or $i=j$;
\item $\sum_{j=1}^nP_{ij}=\sum_{j\in N_i}P_{ij}=1,\forall i\in [n]$; 
\item $\sum_{i=1}^nP_{ij}=\sum_{i\in N_j}P_{ij}=1,\forall j\in [n]$.
\end{compactitem}
Moreover, $P$ is positive semidefinite, and its second largest singular value denoted by $\sigma_2(P)$ is strictly smaller than $1$.
\end{assum}
\begin{assum}
\label{assum4}
At each round $t\in[T]$, the loss function $f_{t,i}(\x)$ of each learner $i\in [n]$ is $G$-Lipschitz over $\K$, i.e., $|f_{t,i}(\x)-f_{t,i}(\mathbf{y})|\leq G\|\x-\mathbf{y}\|_2$ for any $\x,\mathbf{y}\in \K$.
\end{assum}
\begin{assum}
\label{assum1}
The set $\K$ contains the origin, i.e., $\ze\in\K$, and its radius is bounded by $R$, i.e., $\|\x\|_2\leq R$ for any $\x\in \K$.
\end{assum}
\begin{assum}
\label{scvx-assum}
At each round $t\in[T]$, the loss function $f_{t,i}(\x)$ of each learner $i\in [n]$ is $\alpha$-strongly convex over $\K$, i.e., $f_{t,i}(\mathbf{y})\geq f_{t,i}(\x)+\langle\nabla f_{t,i}(\x),\mathbf{y}-\x\rangle+\frac{\alpha}{2}\|\mathbf{y}-\x\|_2^2$ for any $\x,\mathbf{y}\in \K$.
\end{assum}
Note that Assumption \ref{scvx-assum} with $\alpha=0$ reduces to the case with general convex functions.

Then, we briefly introduce the accelerated gossip strategy \citep{Acc_Gossip}, which will be utilized to develop our algorithms. Given a set of vectors denoted as $\nabla_1,\dots,\nabla_n\in\mathbb{R}^d$, a naive idea for approximating the average $\bar{\nabla}=\frac{1}{n}\sum_{i=1}^n\nabla_n$ in the decentralized setting is to perform multiple standard gossip steps \citep{Xiao-Gossip04}, i.e., setting $\nabla_i^{0}=\nabla_i$ and updating as
\begin{equation}
\label{Naiva-Multi-Gossip}
\nabla_i^{k+1}=\sum_{j\in N_i}P_{ij}\nabla_j^{k} \text{~for~} k=0,1,\dots,L-1
\end{equation}
where $L\geq 1$ is the number of iterations. Under Assumption \ref{assum5}, it is well-known that $\nabla_i^{L}$ generated by \eqref{Naiva-Multi-Gossip} provably converges to the average $\bar{\nabla}$ with the increase of $L$. However, \citet{Acc_Gossip} have shown that it is not the most efficient way, and proposed an accelerated gossip strategy by mixing the standard gossip step with an old averaging estimation, i.e., setting $\nabla_i^{0}=\nabla_i^{-1}=\nabla_i$ and updating as
\begin{equation}
\label{Acce-Multi-Gossip}
\nabla_i^{k+1}=(1+\theta)\sum_{j\in N_i}P_{ij}\nabla_j^{k}-\theta \nabla_i^{k-1} \text{~for~} k=0,1,\dots,L-1
\end{equation}
where $\theta>0$ is the mixing coefficient. Let $X^k=\left[(\nabla_i^k)^\top;\dots;(\nabla_n^k)^\top\right]\in\mathbb{R}^{n\times d}$ for any integer $k\geq-1$. For any integer $k\geq 0$, it is not hard to verify that \eqref{Acce-Multi-Gossip} ensures
\begin{equation}
\label{eq-fastMix-pre}
X^{k+1}=(1+\theta)PX^{k}-\theta X^{k-1}.
\end{equation}
This process enjoys the following convergence property, where $\bar{X}=\frac{1}{n}\mathbf{1}\mathbf{1}^\top X^0=\left[\bar{\nabla}^\top;\dots;\bar{\nabla}^\top\right]$ and $\mathbf{1}$ is the all-ones vector in $\mathbb{R}^n$.
\begin{lem}
\label{fastMix-matrix}
(Proposition 1 in \citet{Ye2020}) Under Assumption \ref{assum5}, for $L\geq 1$, the iterations of (\ref{eq-fastMix-pre}) with
$\theta=(1+\sqrt{1-\sigma_2^2(P)})^{-1}$ ensure that
\[\left\|X^L-\bar{X}\right\|_F\leq \sqrt{14}\left(1-\left(1-\frac{1}{\sqrt{2}}\right)\sqrt{1-\sigma_2(P)}\right)^L\left\|X^0-\bar{X}\right\|_F.\]
\end{lem}

\section{Main Results}
In this section, we first present a novel algorithm with improved regret bounds for D-OCO, and establish nearly matching lower bounds. Then, we develop a projection-free variant of our algorithm to efficiently handle complex constraints.

\subsection{A Novel Algorithm with Improved Regret Bounds}
Before introducing our algorithms, we first compare the regret of D-OCO and OCO, which provides insights into our improvements. Specifically, compared with the $O(\sqrt{T})$ regret of OGD and FTRL for OCO, the $O(n^{5/4}\rho^{-1/2}\sqrt{T})$ regret of D-OGD and D-FTRL has an additional factor of $n^{5/4}\rho^{-1/2}$. We notice that this factor reflects the effect of the network size and topology, and is caused by the approximation error of the standard gossip step. For example, a critical part of the analysis for D-FTRL \citep{D-ODA} is the following bound
\begin{equation}
\label{Gossip_Error}
\|\z_i(t)-\bar{\z}(t)\|_2=O\left(\frac{\sqrt{n}}{\rho}\right)
\end{equation}
where $\z_i(t)$ is defined in \eqref{DFTRL}, $\bar{\z}(t)=\frac{1}{n}\sum_{i=1}^n\z_i(t)$ denotes the average $\z_i(t)$ of all learners, and $\rho=1-\sigma_2(P)$. Since $\bar{\z}(t)$ is also equal to $\sum_{\tau=1}^{t-1}\bar{\g}(\tau)$ where $\bar{\g}(\tau)=\frac{1}{n}\sum_{i=1}^n\nabla f_{\tau,i}(\x_{i}(\tau))$, the regret of D-FTRL can be upper bounded by the regret of a virtual centralized update with $\bar{\z}(t)$ plus the cumulative effect of the approximation error in \eqref{Gossip_Error} \citep{D-ODA}, i.e.,
\begin{equation}
\label{D-FTRL-Bound}
R_{T,i}=O\left(\frac{n}{\eta}+n\eta T\right)+O\left(n\eta T\frac{\sqrt{n}}{\rho}\right)=O\left(\frac{n}{\eta}+\frac{n^{3/2}\eta T}{\rho}\right).
\end{equation}
By minimizing the bound in \eqref{D-FTRL-Bound} with $\eta=O(\sqrt{\rho/(\sqrt{n}T)})$, we obtain the $O(n^{5/4}\rho^{-1/2}\sqrt{T})$ regret of D-FTRL.

Thus, to reduce the regret of D-OCO, we should control the approximation error caused by the standard gossip step. Moreover, to handle convex and strongly convex functions with a unified algorithm, there are two possible options:~refining D-OGD or D-FTGL, i.e., the generalized variant of D-FTRL in \eqref{DFTAL}. However, the projection operation in D-OGD makes the analysis of the approximation error more complex. To this end, we propose to improve D-FTGL via the accelerated gossip strategy in \eqref{Acce-Multi-Gossip}. Let $\d_i(t)=\nabla f_{t,i}(\x_i(t))-\alpha\x_i(t)$ and $\bar{\d}(t)=\frac{1}{n}\sum_{i=1}^n\d_i(t)$. According to D-FTGL, we now need to maintain a better $\z_i(t)$ to approximate $\bar{\z}(t)=\sum_{\tau=1}^{t-1}\bar{\d}(\tau)$, which is more general than the above definition. A natural idea is to replace the standard gossip step in \eqref{DFTAL} with multiple accelerated gossip steps, i.e., setting $\z_i(1)=\z_i^{L-1}(1)=\ze$ and computing $\z_i(t)=\z_i^L(t)$ for $t\geq 2$ via the following iterations
\begin{equation}
 \label{naive-A-DFTRL}
\begin{split}
\z_{i}^{k+1}(t)=(1+\theta)\sum_{j\in N_i}P_{ij}\z_j^{k}(t)-\theta\z_i^{k-1}(t) \text{~for~} k=0,1,\dots,L-1
\end{split}
\end{equation}
where $\z_{i}^{0}(t)=\z_{i}(t-1)+\d_i(t-1),~\z_{i}^{-1}(t)=\z_{i}^{L-1}(t-1)+\d_i(t-1)$. One can prove that \eqref{naive-A-DFTRL} ensures (see \eqref{A_key_Tranduction} in Section \ref{sub-sec4.1} for details)
\begin{equation}
\label{temp_equalivace}
\z_i(t)=\z_i^{L}(t)=\sum_{\tau=1}^{t-1}\d_i^{(t-\tau)L}(\tau)
\end{equation}
where $\d_i^{(t-\tau)L}(\tau)$ denotes the output of virtually performing \eqref{Acce-Multi-Gossip} with $\nabla_i=\d_i(\tau)$ and $L=(t-\tau)L$. Due to the convergence behavior of the accelerated gossip strategy, we can control the error of approximating $\bar{\z}(t)$ under any desired level by using a large enough $L$. Unfortunately, this approach requires multiple communications between these learners per round, which is not allowed by D-OCO. 

To address this issue, we design an online accelerated gossip strategy with only one communication per round. The key idea is to incorporate \eqref{naive-A-DFTRL} into a blocking update mechanism \citep{Garber19,Wan-22-JMLR}.
\begin{algorithm}[t]
\caption{AD-FTGL}
\label{ADOA}
\begin{algorithmic}[1]
\STATE \textbf{Input:} $\alpha$, $h$, $\theta$, $L$ 
\STATE \textbf{Initialization:} set $\mathbf{x}_i(1)=\z_i(1)=\z_i^{L-1}(1)=\ze, \forall i\in [n]$
\FOR{$z=1,\dots,T/L$}
\FOR{each local learner $i\in [n]$}
\STATE If $2\leq z$, set $k=0$, $\z_i^0(z)=\z_i(z-1)+\d_i(z-1)$, $\z_i^{-1}(z)=\z_i^{L-1}(z-1)+\d_i(z-1)$
\FOR{$t=(z-1)L+1,\dots,zL$}
\STATE Play $\x_{i}(z)$ and query $\nabla f_{t,i}(\x_{i}(z))$
\STATE If $2\leq z$, update $\z_i^{k+1}(z)=(1+\theta)\sum_{j\in N_i}P_{ij}\z_j^{k}(z)-\theta \z_i^{k-1}(z)$ and $k=k+1$
\ENDFOR
\STATE Set $\d_i(z)=\sum_{t\in\mathcal{T}_z}(\nabla f_{t,i}(\x_i(z))-\alpha\x_i(z))$, where $\mathcal{T}_z=\{(z-1)L+1,\dots,zL\}$
\STATE If $2\leq z$, set $\z_{i}(z)=\z_i^L(z)$
\STATE Compute $\x_i(z+1)=\argmin_{\x\in\K}\left\langle\z_{i}(z),\x\right\rangle+\frac{(z-1)L\alpha}{2}\|\x\|_2^2+h\|\x\|_2^2$
\ENDFOR
\ENDFOR
\end{algorithmic}
\end{algorithm}
To be precise, we divide the total $T$ rounds into $T/L$ blocks, and only maintain a fixed decision $\x_i(z)$ for each learner $i\in [n]$ in block $z$, where $T/L$ is assumed to be an integer without loss of generality. Let $\mathcal{T}_z=\{(z-1)L+1,\dots,zL\}$ denote all rounds contained in each block $z$. With some abuse of notations, we redefine $\d_i(z)=\sum_{t\in\mathcal{T}_z}(\nabla f_{t,i}(\x_i(z))-\alpha\x_i(z))$ and $\bar{\d}(z)=\frac{1}{n}\sum_{i=1}^n\d_i(z)$ for each block $z$. In this way, we only need to maintain a local variable $\z_i(z)$ to approximate $\bar{\z}(z)=\sum_{\tau=1}^{z-1}\bar{\d}(\tau)$ for each learner $i$ in block $z$. The good news is that now $L$ communications can be utilized to update $\z_i(z)$ per block by uniformly allocating them to every round in the block. As a result, we set $\z_i(1)=\z_i^{L-1}(1)=\ze$, and computing $\z_i(z)=\z_i^L(z)$ for $z\geq 2$ in a way similar to \eqref{naive-A-DFTRL}, i.e., performing the following iterations during block $z$
\begin{equation*}
\begin{split}
\z_{i}^{k+1}(z)=(1+\theta)\sum_{j\in N_i}P_{ij}\z_j^{k}(z)-\theta\z_i^{k-1}(z) \text{~for~} k=0,1,\dots,L-1
\end{split}
\end{equation*}
where $\z_i^0(z)=\z_i(z-1)+\d_i(z-1)$ and $\z_i^{-1}(z)=\z_i^{L-1}(z-1)+\d_i(z-1)$. Then, inspired by D-FTGL in \eqref{DFTAL}, we initialize with $\x_i(1)=\ze$, and set the decision $\x_i(z+1)$ for any $i\in [n]$ and $z\geq 1$  as
\begin{equation}
\label{objective-AD-FTGL}
\x_i(z+1) = \argmin_{\x\in\K}\left\langle\z_{i}(z),\x\right\rangle+\frac{(z-1)L\alpha}{2}\|\x\|_2^2+h\|\x\|_2^2.
\end{equation}
We name the proposed algorithm as accelerated
decentralized follow-the-generalized-leader (AD-FTGL), and summarize the complete procedure in Algorithm \ref{ADOA}.

In the following, we first present a lemma regarding the approximation error $\|\z_i(z)-\bar{\z}(z)\|_2$ of AD-FTGL, which demonstrates the advantage of utilizing the accelerated gossip strategy.
\begin{lem}
\label{lem-fastMix}
Let $\bar{\z}(z)=\sum_{\tau=1}^{z-1}\bar{\d}(\tau)$, where $\bar{\d}(\tau)=\frac{1}{n}\sum_{i=1}^n\d_i(\tau)$ and
\begin{equation}
\label{parameters-set}
\theta=\frac{1}{1+\sqrt{1-\sigma_2^2(P)}},~L=\left\lceil\frac{\sqrt{2}\ln(\sqrt{14n})}{(\sqrt{2}-1)\sqrt{1-\sigma_2(P)}}\right\rceil.
\end{equation}
Under Assumptions \ref{assum5}, \ref{assum4}, \ref{assum1}, and \ref{scvx-assum}, for any $i\in [n]$ and $z\in[T/L]$, Algorithm \ref{ADOA} with $\theta,L$ defined in \eqref{parameters-set} ensures \[\left\|\z_i(z)-\bar{\z}(z)\right\|_2\leq 3L(G+\alpha R).\]
\end{lem}
From Lemma \ref{lem-fastMix}, our AD-FTGL can enjoy an error bound of ${O}(\rho^{-1/2}\log n)$ for approximating $\bar{\z}(z)$, which is tighter than the ${O}(\rho^{-1}\sqrt{n})$ error bound in \eqref{Gossip_Error}. By exploiting this improvement, we establish the following guarantee on the regret bound of AD-FTGL.
\begin{thm}
\label{sconvex_upper}
Under Assumptions \ref{assum5}, \ref{assum4}, \ref{assum1}, and \ref{scvx-assum}, for any $i\in [n]$, Algorithm \ref{ADOA} with $\theta,L$ defined in \eqref{parameters-set} ensures
\begin{equation}
\label{unified-thm}
R_{T,i}\leq 3nLG\left(\sum_{z=2}^{T/L}\frac{3L(G+\alpha R)}{(z-2)L\alpha+2h}+\sum_{z=1}^{T/L}\frac{4L(G+2\alpha R)}{zL\alpha+2h}\right)+nhR^2.
\end{equation}
\end{thm}
Then, by combining Theorem \ref{sconvex_upper} with suitable $\alpha$ and $h$, we can achieve specific regret bounds for convex and strongly convex functions, respectively.
\begin{cor}
\label{corollary-convex_upper}
Under Assumptions \ref{assum5}, \ref{assum4}, \ref{assum1}, and \ref{scvx-assum} with $\alpha=0$, for any $i\in [n]$, Algorithm \ref{ADOA} with $\alpha=0$, $h=\sqrt{11LT}G/R$, and $\theta,L$ defined in \eqref{parameters-set} ensures
\begin{equation*}
R_{T,i}\leq 2nGR\sqrt{11LT}.
\end{equation*}
\end{cor}
\begin{cor}
\label{corollary-sconvex_upper}
Under Assumptions \ref{assum5}, \ref{assum4}, \ref{assum1}, and \ref{scvx-assum} with $\alpha>0$, for any $i\in [n]$, Algorithm \ref{ADOA} with $\alpha>0$, $h=\alpha L$, and $\theta,L$ defined in \eqref{parameters-set} ensures
\begin{equation*}
R_{T,i}\leq \frac{3nLG(7G+11\alpha R)(1+\ln(T/L))}{\alpha}+n\alpha LR^2. 
\end{equation*}
\end{cor}
Corollary \ref{corollary-convex_upper} shows that all the learners of AD-FTGL enjoys a regret bound of ${O}(n\rho^{-1/4}\sqrt{T\log n})$ for convex functions, which is tighter than the existing $O(n^{5/4}\rho^{-1/2}\sqrt{T})$ regret bound \citep{DAOL_TKDE,D-ODA} in terms of both $n$ and $\rho$. From Corollary \ref{corollary-sconvex_upper}, all the learners of AD-FTGL can exploit the strong convexity of functions to achieve an ${O}(n\rho^{-1/2}(\log n)\log T)$ regret bound, which has a much tighter dependence on $T$ than the regret bound established by only using the convexity condition. Moreover, it is better than the existing ${O}(n^{3/2}\rho^{-1}\log T)$ regret bound for strongly convex functions \citep{DAOL_TKDE} in terms of both $n$ and $\rho$. 

\begin{remark}
\emph{
One may notice that the values of $\theta$ and $L$ are carefully selected to establish the theoretical guarantees of our AD-FTGL. To emphasize their significance, we further consider two extreme cases: one with $\theta=0$ and the other with $L=1$. First, by setting $\theta=0$, our AD-FTGL is equivalent to improving D-FTGL by only using multiple standard gossip steps. It is easy to verify that due to the slower convergence of standard gossip steps \citep{Xiao-Gossip04}, this extreme case requires a larger $L={O}(\rho^{-1}\log n)$ to achieve an even worse error bound of ${O}(\rho^{-1}\log n)$ for approximating $\bar{\z}(z)$.~Correspondingly, the regret bounds of AD-FTGL for convex and strongly convex functions will degenerate to ${O}(n\rho^{-1/2}\sqrt{T\log n})$ and ${O}(n\rho^{-1}(\log n)\log T)$. Although these bounds are still tighter than the existing $O(n^{5/4}\rho^{-1/2}\sqrt{T})$ and ${O}(n^{3/2}\rho^{-1}\log T)$ regret bounds respectively, their dependence on $\rho$ is worse than the regret bounds achieved in Corollaries \ref{corollary-convex_upper} and \ref{corollary-sconvex_upper}. Moreover, by revisiting the analysis of D-FTGL \citep{Wan-22-JMLR}, we show that the approximation error of its standard gossip step can be improved to $O(\rho^{-1}\log n)$, instead of $O(\rho^{-1}\sqrt{n})$ in \eqref{Gossip_Error} (see Appendix \ref{appendix_last} for details). This result allows us to establish the ${O}(n\rho^{-1/2}\sqrt{T\log n})$ and ${O}(n\rho^{-1}(\log n)\log T)$ regret bounds for convex and strongly convex functions via the original D-FTGL. In other words, it is unnecessary to combine the blocking update mechanism with the standard gossip step. In contrast, the blocking update mechanism is critical for exploiting the accelerated gossip strategy. Note that if $L=1$, our AD-FTGL becomes a non-blocked combination of D-FTGL with the accelerated gossip strategy. Following previous studies on D-OCO \citep{DAOL_TKDE,D-ODA}, such a non-blocked combination may be more natural than the blocked version. However, in this way, the distance between $\z_i(z)$, which satisfies \eqref{temp_equalivace} with $t=z$ and $L=1$, and $\bar{\z}(z)$ cannot be controlled as desired. Specifically, due to the newly added component in $\z_i(z)$, i.e., $\d_i^{z-\tau}(\tau)$ in \eqref{temp_equalivace} for $\tau$ close to $z-1$, we can only modify the analysis of Lemma \ref{lem-fastMix} to derive a worse error bound of ${O}(\rho^{-1/2}\sqrt{n})$. Correspondingly, the regret bounds of AD-FTGL for convex and strongly convex functions will degenerate to ${O}(n^{5/4}\rho^{-1/4}\sqrt{T})$ and ${O}(n^{3/2}\rho^{-1/2}\log T)$, whose dependence on $n$ is much worse than the regret bounds achieved in Corollaries \ref{corollary-convex_upper} and \ref{corollary-sconvex_upper}.
}
\end{remark}

\subsection{Lower Bounds}
Although there still exist gaps between our improved regret bounds and the $\Omega(n\sqrt{T})$ and $\Omega(n)$ lower bounds established by \citet{Wan-22-JMLR}, this is mainly because they do not take the decentralized structure into account. To fill these gaps, we maximize the hardness of D-OCO by considering the $1$-connected cycle graph \citep{DADO2011}, i.e., constructing the graph $\mathcal{G}$ by placing the $n$ nodes on a circle and only connecting each node to one neighbor on its right and left. In this topology, the adversary can make at least one learner, e.g., learner 1, suffer $O(n)$ communication delays for receiving the information of the global function $f_t(\x)$. Because of the $O(n)$ communication delays, we can establish $\Omega(n\sqrt{nT})$ and $\Omega(n^2)$ lower bounds for convex and strongly convex functions, respectively. Then, by exploiting the dependence of spectral properties on the network size $n$, we obtain lower bounds involving the spectral gap. Moreover, inspired by \citet{Wan-22-JMLR}, we also make an extension to obtain the following lower bounds for the more challenging setting with only $C$ communication rounds.
\begin{thm}
\label{thm_lowerB}
Suppose $\K=[-R/\sqrt{d},R/\sqrt{d}]^d$ which satisfies Assumption \ref{assum1}, and $n=2(m+1)$ for some positive integer $m$.  For any D-OCO algorithm communicating $C$ rounds before round $T$, there exists a sequence of loss functions satisfying Assumption \ref{assum4}, a graph $\mathcal{G}=([n],E)$, and a matrix $P$ satisfying Assumption \ref{assum5} such that
\[
\text{~if~} n\leq 8C+16, R_{T,1}\geq\frac{n\sqrt{\pi}RGT}{16(1-\sigma_2(P))^{1/4}\sqrt{C+1}}  \text{,~and~otherwise,~} R_{T,1}\geq \frac{nRGT}{4}.
\]
\end{thm}
\begin{thm}
\label{thm_lowerB-sc}
Suppose $\K=[-R/\sqrt{d},R/\sqrt{d}]^d$ which satisfies Assumption \ref{assum1}, and $n=2(m+1)$ for some positive integer $m$. For any D-OCO algorithm communicating $C$ rounds before round $T$, there exists a sequence of loss functions satisfying Assumption \ref{scvx-assum} and Assumption \ref{assum4} with $G=2\alpha R$, a graph $\mathcal{G}=([n],E)$, and a matrix $P$ satisfying Assumption \ref{assum5} such that
\[
\text{~if~} n\leq 8C+16, R_{T,1}\geq\frac{\alpha \pi n R^2T}{256(C+1)\sqrt{1-\sigma_2(P)}}  \text{,~and~otherwise,~} R_{T,1}\geq \frac{\alpha nR^2T}{16}.
\]
\end{thm}
Note that in previous studies \citep{DAOL_TKDE,D-ODA} and this paper, the upper regret bounds of D-OCO algorithms generally hold for all possible graphs and communication matrices $P$ satisfying Assumption \ref{assum5}. Therefore, although lower bounds in our Theorems \ref{thm_lowerB} and \ref{thm_lowerB-sc} only hold for a specific choice of the graph and $P$, they are sufficient to prove the tightness of the upper bound in general. Specifically, by combining Theorem \ref{thm_lowerB} with $C=O(T)$, we can establish a lower bound of $\Omega(n\rho^{-1/4}\sqrt{T})$ for D-OCO with convex functions, which matches the ${O}(n\rho^{-1/4}\sqrt{T\log n})$ regret of our AD-FTGL up to polylogarithmic factors in $n$. For D-OCO with strongly convex functions, a lower bound of $\Omega(n\rho^{-1/2})$ can be established by combining Theorem \ref{thm_lowerB-sc} with $C=O(T)$, which matches the ${O}(n\rho^{-1/2}(\log n)\log T)$ regret of our AD-FTGL up to polylogarithmic factors in both $n$ and $T$. 

Now, our AD-FTGL have been shown to be nearly optimal for D-OCO with both convex and strongly convex functions. Nonetheless, there still exists an unsatisfactory point in the lower bound for strongly convex functions, as it cannot recover the well-known $\Omega(\log T)$ lower bound for OCO with strongly convex functions \citep{Abernethy08,Epoch-GD}. To address this issue, we establish the following result by extending the analysis of \citet{Epoch-GD} from OCO into D-OCO.
\begin{thm}
\label{imp-thm_lowerB-sc}
Suppose $\K=[0,R/\sqrt{d}]^d$, which satisfies Assumption \ref{assum1} and $n=2(m+1)$ for some positive integer $m$. For any D-OCO algorithm, if $16n+1\leq T$, there exists a sequence loss functions satisfying Assumption \ref{scvx-assum} and Assumption \ref{assum4} with $G=\alpha R$, a graph $\mathcal{G}=([n],E)$, and a matrix $P$ satisfying Assumption \ref{assum5} such that
\[
R_{T,1}\geq\frac{16^{-5}\alpha\pi (\log_{16}(30(T-1)/n)-2)(n-2)R^2}{4\sqrt{1-\sigma_2(P)}}.
\]
\end{thm}
Compared with Theorem \ref{thm_lowerB-sc}, Theorem \ref{imp-thm_lowerB-sc} establishes an improved lower bound of $\Omega(n\rho^{-1/2}\log T)$ for D-OCO with strongly convex functions, which matches the ${O}(n\rho^{-1/2}(\log n)\log T)$ regret of our AD-FTGL up to polylogarithmic factors in only $n$. 
\begin{remark}
\emph{
One may also wonder whether it is possible to extend the result in Theorem \ref{imp-thm_lowerB-sc} into the setting with only $C$ communication rounds. However, there do exist some technical challenges for this extension (see discussions at the end of the proof of Theorem \ref{imp-thm_lowerB-sc} for details), and thus we leave it as a future work.
}
\end{remark}
\begin{algorithm}[t]
\caption{CG}
\label{CG}
\begin{algorithmic}[1]
\STATE \textbf{Input:} $\K$, $K$, $F(\x)$, $\x_{\ii}$
\STATE \textbf{Initialization:} $\y_0=\x_{\ii}$
\FOR{$k=0,\dots,K-1$}
\STATE $\vv_k=\argmin_{\x\in\K}\langle\nabla F(\y_k),\x\rangle$
\STATE $s_k=\argmin_{s\in[0,1]}F(\y_{k}+s(\vv_k-\y_{k}))$
\STATE $\y_{k+1}=\y_{k}+s_k(\vv_k-\y_{k})$
\ENDFOR
\STATE \textbf{return} $\x_{\oo}=\y_{K}$
\end{algorithmic}
\end{algorithm}
\subsection{A Projection-free Variant of Our Algorithm}
Furthermore, to efficiently handle applications with complex constraints, we propose a projection-free variant of our AD-FTGL. Following the existing projection-free D-OCO algorithm in \citet{Wan-22-JMLR}, our main idea is to combine AD-FTGL with conditional gradient (CG)---a classical projection-free algorithm for offline optimization \citep{FW-56,Revist_FW}. Specifically, the detailed procedure of CG is outlined in Algorithm \ref{CG}. Given a function $F(\x):\K\mapsto\mathbb{R}$ and an initial point $\y_0=\x_{\ii}\in\K$, it iteratively performs $K$ linear optimization steps as shown from steps $3$ to $7$, and finally outputs $\x_{\oo}=\y_{K}$. To make AD-FTGL projection-free, it is natural to approximately solve \eqref{objective-AD-FTGL} via CG. 

However, there are still some technical details that require careful attention. First, for every invocation of CG, the number of iterations must equal to the block size, i.e., $K=L$, which ensures that each learner at most requires $T$ linear optimization steps in total. Otherwise, even only using linear optimization steps, the time complexity could be equivalent to that of projection-based algorithms. Second, a straightforward combination of \eqref{objective-AD-FTGL} and CG requires the algorithm to stop at the end of each block and wait until $L$ linear optimization steps are completed. To avoid this issue, we simply set $\x_i(1)=\x_i(2)=\ze$, and then compute $\x_i(z+1)$ based on $\z_i(z-1)$, rather than $\z_i(z)$ used in \eqref{objective-AD-FTGL}, i.e., computing
\begin{equation*}
\x_i(z+1)=\text{CG}(\K,L,F_{z,i}(\x),\x_i(z))
\end{equation*}
where $F_{z,i}(\x)$ denotes the intermediate objective function based on $\z_i(z-1)$, i.e.,
\[F_{z,i}(\x)=\left\langle\z_{i}(z-1),\x\right\rangle+\frac{(z-2)L\alpha}{2}\|\x\|_2^2+h\|\x\|_2^2.\] In this way, the $L$ linear optimization steps required by CG can be uniformly allocated to every round in block $z$, since $\z_{i}(z-1)$ is available at the beginning of this block. Last, inspired by \citet{Wan-22-JMLR}, to control the approximation error of CG, $L$ now should be much larger than that defined in \eqref{parameters-set}. Note that the latter is sufficient for generating a good approximation of $\bar{\z}(z)$. Thus, we keep the number of accelerated gossip steps used in each block unchanged, and denote the specific value by $L^\prime$. This enables us to achieve sublinear communication complexity.
\begin{algorithm}[t]
\caption{Projection-free Variant of AD-FTGL}
\label{Projection-Free}
\begin{algorithmic}[1]
\STATE \textbf{Input:} $\alpha$, $h$, $\theta$, $L$, $L^\prime$ 
\STATE \textbf{Initialization:} set $\mathbf{x}_i(1)=\x_i(2)=\z_i(1)=\z_i^{L^\prime-1}(1)=\ze, \forall i\in [n]$
\FOR{$z=1,\dots,T/L$}
\FOR{each local learner $i\in [n]$}
\STATE Define $F_{z,i}(\x)=\left\langle\z_{i}(z-1),\x\right\rangle+\frac{(z-2)L\alpha}{2}\|\x\|_2^2+h\|\x\|_2^2$
\STATE If $2\leq z$, set $k=0$, $\z_i^0(z)=\z_i(z-1)+\d_i(z-1)$, $\z_i^{-1}(z)=\z_i^{L^\prime-1}(z-1)+\d_i(z-1)$
\FOR{$t=(z-1)L+1,\dots,zL$}
\STATE Play $\x_{i}(z)$ and query $\nabla f_{t,i}(\x_{i}(z))$
\STATE If $2\leq z$ and $k< L^\prime$, update $\z_i^{k+1}(z)=(1+\theta)\sum_{j\in N_i}P_{ij}\z_j^{k}(z)-\theta \z_i^{k-1}(z)$ and $k=k+1$
\ENDFOR
\STATE Set $\d_i(z)=\sum_{t\in\mathcal{T}_z}(\nabla f_{t,i}(\x_i(z))-\alpha\x_i(z))$, where $\mathcal{T}_z=\{(z-1)L+1,\dots,zL\}$
\STATE If $2\leq z$, set $\z_{i}(z)=\z_i^{L^\prime}(z)$
\STATE If $2\leq z$, compute $\x_i(z+1)=\text{CG}(\K,L,F_{z,i}(\x),\x_i(z))$
\ENDFOR
\ENDFOR
\end{algorithmic}
\end{algorithm}

The complete procedure of our projection-free algorithm is summarized in Algorithm \ref{Projection-Free}. By combining our original analysis of AD-FTGL with the convergence property of CG, we first establish the following guarantee for the regret bound of our projection-free algorithm.
\begin{thm}
\label{upper-projection-free}
Under Assumptions \ref{assum5}, \ref{assum4}, \ref{assum1}, and \ref{scvx-assum}, for any $i\in [n]$, Algorithm \ref{Projection-Free} with 
\begin{equation}
\label{parameters-set2}
\theta=\frac{1}{1+\sqrt{1-\sigma_2^2(P)}},~L^\prime=\left\lceil\frac{\sqrt{2}\ln(\sqrt{14n})}{(\sqrt{2}-1)\sqrt{1-\sigma_2(P)}}\right\rceil
\end{equation}
ensures
\begin{equation}
\label{unified-thm-pro}
R_{T,i}\leq 3nLG\left(\sum_{z=3}^{T/L}\frac{3L(G+\alpha R)}{(z-3)L\alpha +2h}+\sum_{z=1}^{T/L}\frac{6L(G+2\alpha R)}{(z-1)L\alpha+2h}\right)+nhR^2+\frac{12nGRT}{\sqrt{L+2}}.
\end{equation}
\end{thm}
Then, by combining Theorem \ref{upper-projection-free} with suitable $\alpha$, $h$, and $L$, we can establish specific regret bounds for our projection-free algorithm.
\begin{cor}
\label{corollary-convex_upper-projection-free}
Suppose Assumptions \ref{assum5}, \ref{assum4}, \ref{assum1}, and \ref{scvx-assum} with $\alpha=0$ hold. For any $i\in [n]$, Algorithm \ref{Projection-Free} with $\alpha=0$, $h=\sqrt{14LT}G/R$, $L=\sqrt{T}L^\prime$, and $\theta,L^\prime$ defined in \eqref{parameters-set2} ensures
\begin{equation}
\label{cor3-1}
R_{T,i}\leq 2\sqrt{14}nGR\sqrt{L^\prime}T^{3/4}+\frac{12nGRT^{3/4}}{\sqrt{L^\prime}}. 
\end{equation}
Moreover, if \eqref{parameters-set2} satisfies $L^\prime\leq \sqrt{T}$, for any $i\in [n]$, Algorithm \ref{Projection-Free} with $\alpha=0$, $h=\sqrt{14LT}G/R$, $L=\sqrt{T}$, and $\theta,L^\prime$ defined in \eqref{parameters-set2} ensures
\begin{equation}
\label{cor3-2}
R_{T,i}\leq (2\sqrt{14}+12)nGRT^{3/4}. 
\end{equation}
\end{cor}
\begin{cor}
\label{corollary-sconvex_upper-projection-free}
Suppose Assumptions \ref{assum5}, \ref{assum4}, \ref{assum1}, and \ref{scvx-assum} with $\alpha>0$ hold. For any $i\in [n]$, Algorithm \ref{Projection-Free} with $\alpha>0$, $h=\alpha L$, $L=T^{2/3}(\ln T)^{-2/3}L^\prime$, and $\theta,L^\prime$ defined in \eqref{parameters-set2} ensures
\begin{equation}
\label{cor4-1}
\begin{split}
R_{T,i}\leq&\frac{3nG(9G+15\alpha R)T^{2/3}L^\prime((\ln T)^{-2/3}+(\ln T)^{1/3})}{\alpha}\\
&+n\alpha T^{2/3}L^\prime R^2(\ln T)^{-2/3}+\frac{12nGRT^{2/3}(\ln T)^{1/3}}{\sqrt{L^\prime}}.
\end{split}
\end{equation}
Moreover, if \eqref{parameters-set2} satisfies $L^\prime\leq T^{2/3}(\ln T)^{-2/3}$, for any $i\in [n]$, Algorithm \ref{Projection-Free} with $\alpha>0$, $h=\alpha L$, $L=T^{2/3}(\ln T)^{-2/3}$, and $\theta,L^\prime$ defined in \eqref{parameters-set2} ensures
\begin{equation}
\label{cor4-2}
\begin{split}
R_{T,i}
\leq&\frac{3nG(9G+15\alpha R)T^{2/3}((\ln T)^{-2/3}+(\ln T)^{1/3})}{\alpha}\\
&+n\alpha T^{2/3} R^2(\ln T)^{-2/3}+12nGRT^{2/3}(\ln T)^{1/3}.
\end{split}
\end{equation}
\end{cor}
It is easy to verify that the number of communication rounds used in this algorithm is $TL^{\prime}/L$. Both Corollaries \ref{corollary-convex_upper-projection-free} and \ref{corollary-sconvex_upper-projection-free} present two  parameter choices for our projection-free algorithm, each determined by different trade-offs between the regret and communication complexity. To be precise, from \eqref{cor3-1} and \eqref{cor4-1}, our projection-free algorithm can enjoy an $O(n\rho^{-1/4}\sqrt{\log n}T^{3/4})$ regret bound for convex functions with $O(\sqrt{T})$ communication rounds, and an $O(n\rho^{-1/2}T^{2/3}(\log T)^{1/3}\log n)$ regret bound for strongly convex functions with $O(T^{1/3}(\log T)^{2/3})$ communication rounds. In contrast, with the same number of communication rounds, the existing projection-free algorithm in \citet{Wan-22-JMLR} only achieves the worse $O(n^{5/4}\rho^{-1/2}T^{3/4})$ and ${O}(n^{3/2}\rho^{-1}T^{2/3}(\log T)^{1/3})$ regret bounds for convex functions and strongly convex functions, respectively. Since their projection-free algorithm is a variant of D-FTGL, this comparison implies that our projection-free algorithm can simply inherit the improvement of AD-FTGL over D-FTGL. Moreover, from \eqref{cor3-2} and \eqref{cor4-2}, our projection-free algorithm can further reduce regret bounds for convex and strongly convex functions to $O(nT^{3/4})$ and $O(nT^{2/3}(\log T)^{1/3})$ by increasing the number of communication rounds to $O(\rho^{-1/2}\sqrt{T}\log n)$ and $O(\rho^{-1/2}T^{1/3}(\log T)^{2/3}\log n)$, respectively. This is somewhat surprising since without other additional assumptions, even running existing centralized projection-free OCO algorithms \citep{Hazan2012,SC_OFW} over the global function $f_t(\x)$ can only achieve the same regret bounds (up to polylogarithmic factors in $T$ for strongly convex functions). Finally, according to the lower bounds in Theorems \ref{thm_lowerB} and \ref{thm_lowerB-sc}, the number of communication rounds required by our projection-free algorithm to achieve the above regret bounds is optimal up to polylogarithmic factors in $n$ for convex functions, and polylogarithmic factors in $n$ and $T$ for strongly convex functions, respectively.
\section{Theoretical Analysis}
Here, we provide the proofs of our theoretical guarantees on AD-FTGL, lower bounds, and the projection-free variant of AD-FTGL. The refined analysis for D-FTGL can be found in the appendix.
\subsection{Proof of Lemma \ref{lem-fastMix}}
\label{sub-sec4.1}
Let $\d_i^0(z)=\d_i^{-1}(z)=\d_i(z)$. For any $i\in [n]$, $z\in[T/L-1]$, and any non-negative integer $k$, we first define a virtual update as
\begin{equation}
\label{temp0}
\d_i^{k+1}(z)=(1+\theta)\sum_{j\in N_i}P_{ij}\d_j^{k}(z)-\theta\d_i^{k-1}(z).
\end{equation}
In the following, we will prove that for any $z=2,\dots,T/L$, Algorithm \ref{ADOA} ensures
\begin{equation}
\label{A_key_Tranduction}
\z_i^k(z)=\sum_{\tau=1}^{z-1}\d_i^{(z-\tau-1)L+k}(\tau),~\forall k=1,\dots,L
\end{equation}
by the induction method.

For $z=2$, it is easy to verify that \eqref{A_key_Tranduction} holds due to $\z_i^0(2)=\z_i^{-1}(2)=\d_i(1)$ and \eqref{temp0}. Then, we assume that \eqref{A_key_Tranduction} holds for some $z>2$, and prove it also holds for $z+1$. From step $5$ of our Algorithm \ref{ADOA}, we have
\begin{equation}
\label{propert1}
\begin{split}
&\z_i^{0}(z+1)=\z_i(z)+\d_i(z)=\z_i^L(z)+\d_i^0(z)\overset{\eqref{A_key_Tranduction}}{=}\sum_{\tau=1}^{z}\d_i^{(z-\tau)L}(\tau),\\
&\z_i^{-1}(z+1)=\z_i^{L-1}(z)+\d_i(z)=\z_i^{L-1}(z)+\d_i^{-1}(z)\overset{\eqref{A_key_Tranduction}}{=}\sum_{\tau=1}^{z}\d_i^{(z-\tau)L-1}(\tau).
\end{split}
\end{equation}
By combining \eqref{propert1} with step $8$ of Algorithm \ref{ADOA}, for $k=1$, we have
\begin{equation}
\label{divide-for-pre1}
\begin{split}
\z_i^{k}(z+1)=&(1+\theta)\sum_{j\in N_i}P_{ij}\z_j^{k-1}(z+1)-\theta \z_i^{k-2}(z+1)\\
=&(1+\theta)\sum_{j\in N_i}P_{ij}\sum_{\tau=1}^{z}\d_j^{(z-\tau)L+k-1}(\tau)-\theta \sum_{\tau=1}^{z}\d_i^{(z-\tau)L-1+k-1}(\tau).
\end{split}
\end{equation}
By reorganizing the right side of \eqref{divide-for-pre1}, for $k=1$, we have
\begin{equation}
\label{divide-for-pre2}
\begin{split}
\z_i^{k}(z+1)
=&\sum_{\tau=1}^{z}\left((1+\theta)\sum_{j\in N_i}P_{ij}\d_j^{(z-\tau)L+k-1}(\tau)-\theta \d_i^{(z-\tau)L-1+k-1}(\tau)\right)\\
\overset{\eqref{temp0}}{=}&\sum_{\tau=1}^{z}\d_i^{(z-\tau)L+k}(\tau).
\end{split}
\end{equation}
By repeating \eqref{divide-for-pre1} and \eqref{divide-for-pre2}  for $k=2,\dots,L$, the proof of \eqref{A_key_Tranduction} for $z+1$ is completed.

Then, from \eqref{A_key_Tranduction}, for any $i\in[n]$ and $z=2,\dots,T/L$, we have
\begin{equation}
\label{temp2}
\left\|\z_i(z)-\bar{\z}(z)\right\|_2=\left\|\sum_{\tau=1}^{z-1}\d_i^{(z-\tau)L}(\tau)-\sum_{\tau=1}^{z-1}\bar{\d}(\tau)\right\|_2\leq\sum_{\tau=1}^{z-1}\left\|\d_i^{(z-\tau)L}(\tau)-\bar{\d}(\tau)\right\|_2.
\end{equation}
To further analyze the right side of \eqref{temp2}, we define \[X^k=\left[{\d}_{1}^k(\tau)^\top;\dots;{\d}_{n}^k(\tau)^\top\right]\in\mathbb{R}^{n\times d}\] for any integer $k\geq-1$ and $\bar{X}=\frac{1}{n}\mathbf{1}\mathbf{1}^\top X^0=\left[\bar{\d}(\tau)^\top;\dots;\bar{\d}(\tau)^\top\right]$. According to \eqref{temp0}, it is not hard to verify that the sequence of $X^1,\dots,X^L$ follows the update rule in \eqref{eq-fastMix-pre}.

Let $c=1-1/\sqrt{2}$. Then, by combining with Lemma \ref{fastMix-matrix}, for any $\tau\leq z$, we have
\begin{equation}
\label{temp3}
\begin{split}
& \left\|X^{(z-\tau)L}-\bar{X}\right\|_F\leq\sqrt{14}\left(1-c\sqrt{1-\sigma_2(P)}\right)^{(z-\tau)L}\left\|X^0-\bar{X}\right\|_F\\
\leq&\sqrt{14}\left(1-c\sqrt{1-\sigma_2(P)}\right)^{(z-\tau)L}\left(\left\|X^0\right\|_F+\left\|\bar{X}\right\|_F\right)\\
=&\sqrt{14}\left(1-c\sqrt{1-\sigma_2(P)}\right)^{(z-\tau)L}\left(\sqrt{\sum_{i=1}^n\|{\d}_{i}(\tau)\|_2^2}+\sqrt{n\|{\bar{\d}}(\tau)\|_2^2}\right).
\end{split}
\end{equation}
Because of Assumptions \ref{assum4} and \ref{assum1}, for any $z\in[T/L]$ and $i\in [n]$, it easy to verify that
\begin{equation}
\label{ub_eq2_sc}
\begin{split}
&\|\d_i(z)\|_2=\left\|\sum_{t\in\mathcal{T}_{z}}\left(\nabla f_{t,i}(\x_{i}(z))-\alpha\x_{i}(z)\right)\right\|_2\leq L(G+\alpha R),\\
&\|\bar{\d}(z)\|_2=\left\|\frac{1}{n}\sum_{i=1}^n\d_i(z)\right\|_2\leq L(G+\alpha R).
\end{split}
\end{equation}
By combining \eqref{temp3} with \eqref{ub_eq2_sc}, for any $i\in[n]$ and $\tau\leq z$, we have
\begin{equation}
\label{temp4}
\begin{split}
\left\|\d_i^{(z-\tau)L}(\tau)-\bar{\d}(\tau)\right\|_2\leq&\left\|X^{(z-\tau)L}-\bar{X}\right\|_F\\
\leq&2\sqrt{14n}\left(1-c\sqrt{1-\sigma_2(P)}\right)^{(z-\tau)L}L(G+\alpha R).
\end{split}
\end{equation}
Moreover, because of the value of $L$ in \eqref{parameters-set}, we have
\begin{equation}
\label{epsilon_upper}
\begin{split}
\epsilon=&\left(1-c\sqrt{1-\sigma_2(P)}\right)^L\leq\left(1-c\sqrt{1-\sigma_2(P)}\right)^{\frac{\ln(\sqrt{14n})}{c\sqrt{1-\sigma_2(P)}}}\\
\leq&\left(1-c\sqrt{1-\sigma_2(P)}\right)^{\frac{\ln(\sqrt{14n})}{\ln\left(1-c\sqrt{1-\sigma_2(P)}\right)^{-1}}}= \frac{1}{\sqrt{14n}}
\end{split}
\end{equation}
where the second inequality is due to $\ln x^{-1}\geq 1-x$ for any $x>0$.

By combining \eqref{temp2} with \eqref{temp4} and \eqref{epsilon_upper}, for any $i\in[n]$ and $z=2,\dots,T/L$, we have
\begin{equation}
\label{temp2-1}
\begin{split}
\left\|\z_i(z)-\bar{\z}(z)\right\|_2\leq&2L(G+\alpha R)\sqrt{14n}\sum_{\tau=1}^{z-1}\epsilon^{(z-\tau)}\overset{\eqref{epsilon_upper}}{\leq} 2L(G+\alpha R)\sum_{\tau=1}^{z-1}\epsilon^{(z-\tau-1)}\\
\leq&\frac{2L(G+\alpha R)}{1-\epsilon}\overset{\eqref{epsilon_upper}}{\leq}2L(G+\alpha R)+\frac{2L(G+\alpha R)}{\sqrt{14n}-1}\leq 3L(G+\alpha R)
\end{split}
\end{equation}
where the last inequality is due to $\sqrt{14n}>3$ for any $n\geq 1$. Now, we can complete the proof of Lemma \ref{lem-fastMix} by combining \eqref{temp2-1} with $\|\z_i(1)-\bar{\z}(1)\|_2=0$.

\subsection{Proof of Theorem \ref{sconvex_upper}}
According to Algorithm \ref{ADOA}, the total $T$ rounds are divided into $T/L$ blocks. For $z\in[T/L+1]$, we define a virtual decision
\begin{equation}
\label{virtual_decision_scvx}
\bar{\x}(z)=\argmin_{\x\in\K}\left\langle\x,\bar{\z}(z)\right\rangle+\frac{(z-1)L\alpha}{2}\|\x\|_2^2+h\|\x\|_2^2
\end{equation}
where $\bar{\z}(z)=\sum_{\tau=1}^{z-1}\bar{\d}(\tau)$ and $\bar{\d}(\tau)=\frac{1}{n}\sum_{i=1}^n\d_i(\tau)$. In the following, we will bound the regret of any learner $i$ by analyzing the regret of $\bar{\x}(2),\dots,\bar{\x}(T/L+1)$ on a sequence of loss functions defined by $\bar{\d}(1),\dots,\bar{\d}(T/L)$ and the distance $\|\x_i(z)-\bar{\x}(z+1)\|_2$ for any $z\in[T/L]$. To this end, we first introduce two useful lemmas.
\begin{lem}
\label{lem-ftl}
(Lemma 6.6 in \citet{Garber16}) Let $\{\ell_t(\x)\}_{t=1}^T$ be a sequence of functions and $\x_t^\ast\in\argmin_{\x\in\K}\sum_{\tau=1}^t\ell_{\tau}(\x)$ for any $t\in[T]$. Then, it holds that
\[\sum_{t=1}^T\ell_t(\x_t^\ast)-\min_{\x\in\K}\sum_{t=1}^T\ell_t(\x)\leq 0.\]
\end{lem}
\begin{lem}
\label{lem-stab}
(Lemma 5 in \citet{DADO2011}) Let $\Pi_\K(\uu,\eta)=\argmin_{\x\in\K}\langle\uu,\x\rangle+\frac{1}{\eta}\|\x\|_2^2$. For any $\uu,\vv\in\mathbb{R}^d$, we have
\[\|\Pi_\K(\uu,\eta)-\Pi_\K(\vv,\eta)\|_2\leq\frac{\eta}{2}\|\uu-\vv\|_2.\]
\end{lem}
Let $\ell_z(\x)=\langle \x,\bar{\d}(z)\rangle+\frac{L\alpha}{2}\|\x\|_2^2$ for any $z\in[T/L]$. Combining Lemma \ref{lem-ftl} with \eqref{virtual_decision_scvx}, for any $\x\in\K$, it is easy to verify that
\begin{equation}
\label{sub_eq1}
\begin{split}
\sum_{z=1}^{T/L}\ell_z(\bar{\x}(z+1))-\sum_{z=1}^{T/L}\ell_z(\x)\leq h\left(\|\x\|_2^2-\|\bar{\x}(2)\|_2^2\right)\leq hR^2
\end{split}
\end{equation}
where the last inequality is due to Assumption \ref{assum1} and $\|\bar{\x}(2)\|_2^2\geq0$.

Then, we also notice that Algorithm \ref{ADOA} ensures
\begin{equation}
\label{pract_decision_scvx}
\x_i(z)=\argmin_{\x\in\K}\left\langle\x, \z_i(z-1)\right\rangle+\frac{(z-2)L\alpha}{2}\|\x\|_2^2+h\|\x\|_2^2.
\end{equation}
for any $z=2,\dots,T/L$. 

By combining Lemma \ref{lem-stab} with \eqref{virtual_decision_scvx} and \eqref{pract_decision_scvx}, for any $z=2,\dots,T/L$, we have
\begin{equation}
\label{sub_eq4}
\begin{split}
\|\x_i(z)-\bar{\x}(z-1)\|_2\leq&\frac{\left\|\z_i(z-1)-\bar{\z}(z-1)\right\|_2}{(z-2)L\alpha+2h}\leq\frac{3L(G+\alpha R)}{(z-2)L\alpha+2h}
\end{split}
\end{equation}
where the last inequality is due to Lemma \ref{lem-fastMix}.

To bound $\|\x_i(z)-\bar{\x}(z+1)\|_2$, we still need to analyze the term $\|\bar{\x}(z)-\bar{\x}(z+1)\|_2$ for any $z\in[T/L]$. Let $F_z(\x)=\sum_{\tau=1}^{z}\ell_{\tau}(\x)+h\|\x\|_2^2$ for any $z\in[T/L]$. It is easy to verify that $F_z(\x)$
is $(zL\alpha+2h)$-strongly convex over $\K$, and $\bar{\x}(z+1)=\argmin_{\x\in\K} F_z(\x)$.
Note that as proved by \citet{Hazan2012}, for any $\alpha$-strongly convex function $f(\x):\K\mapsto\mathbb{R}$ and $\x\in\K$, it holds that
\begin{equation}
\label{sub_eq5}
\frac{\alpha}{2}\|\x-\x^\ast\|_2^2\leq f(\x)-f(\x^\ast)
\end{equation}
where $\x^\ast=\argmin_{\x\in\K}f(\x)$. Moreover, for any $\x,\y\in\K$ and $z\in[T/L]$, we have
\begin{equation}
\label{sub_eq6}
\begin{split}
|\ell_z(\x)-\ell_z(\y)|\leq&\left|\langle \nabla \ell_z(\x),\x-\y\rangle\right|\leq\|\nabla \ell_z(\x)\|_2\|\x-\y\|_2\\
=&\|\bar{\d}(z)+\alpha L\x\|_2\|\x-\y\|_2\overset{\eqref{ub_eq2_sc}}{\leq} L(G+2\alpha R)\|\x-\y\|_2.
\end{split}
\end{equation}
Then, for any $z^\prime\leq z\in[T/L]$, it is not hard to verify that
\begin{equation*}
\begin{split}
&\|\bar{\x}(z^\prime)-\bar{\x}(z+1)\|_2^2\\
\overset{\eqref{sub_eq5}}{\leq}&\frac{2}{zL\alpha+2h}(F_z(\bar{\x}(z^\prime))-F_z(\bar{\x}(z+1)))\\
=&\frac{2}{zL\alpha+2h}\left(F_{z^\prime-1}(\bar{\x}(z^\prime))-F_{z^\prime-1}(\bar{\x}(z+1))+\sum_{\tau=z^\prime}^{z}(\ell_\tau(\bar{\x}(z^\prime))-\ell_\tau(\bar{\x}(z+1)))\right)\\
\overset{\eqref{virtual_decision_scvx}}{\leq}&\frac{2\sum_{\tau=z^\prime}^{z}(\ell_\tau(\bar{\x}(z^\prime))-\ell_\tau(\bar{\x}(z+1)))}{zL\alpha+2h}\\
\overset{\eqref{sub_eq6}}{\leq}&\frac{2(z-z^\prime+1)L(G+2\alpha R)\|\bar{\x}(z^\prime)-\bar{\x}(z+1)\|_2}{zL\alpha+2h}
\end{split}
\end{equation*}
which implies that
\begin{equation}
\begin{split}
\label{sub_eq7}
\|\bar{\x}(z^\prime)-\bar{\x}(z+1)\|_2\leq\frac{2(z-z^\prime+1)L(G+2\alpha R)}{zL\alpha+2h}.
\end{split}
\end{equation}
By combining (\ref{sub_eq4}) and (\ref{sub_eq7}), for any $z=2,\dots,T/L$, we have
\begin{equation}
\label{sub_eq8}
\begin{split}
\|\x_i(z^\prime)-\bar{\x}(z+1)\|_2\leq&\|\x_i(z)-\bar{\x}(z-1)\|_2+ \|\bar{\x}(z-1)-\bar{\x}(z+1)\|_2\\
\leq&\frac{3L(G+\alpha R)}{(z-2)\alpha+2h}+\frac{4L(G+2\alpha R)}{zL\alpha+2h}.
\end{split}
\end{equation}
For $z=1$, we notice that $\x_i(1)=\bar{\x}(1)=\ze$, and it is easy to verify that
\begin{equation}
\label{sub_eq9}
\begin{split}
\|\x_i(z)-\bar{\x}(z+1)\|_2= \|\bar{\x}(z)-\bar{\x}(z+1)\|_2\overset{\eqref{sub_eq7}}{\leq}\frac{2L(G+2\alpha R)}{zL\alpha+2h}.
\end{split}
\end{equation}
For brevity, let $\epsilon_z$ denote the upper bound of $\|\x_i(z)-\bar{\x}(z+1)\|_2$ derived in \eqref{sub_eq8} and \eqref{sub_eq9}.

For any $z\in[T/L]$, $t\in\mathcal{T}_z$, $j\in [n]$, and $\x\in\K$, because of Assumptions \ref{assum4} and \ref{scvx-assum}, we have
\begin{equation}
\label{for_ref-D-FTGL}
\begin{split}
f_{t,j}(\x_i(z))-f_{t,j}(\x)\leq& f_{t,j}(\x_j(z))-f_{t,j}(\x)+G\|\x_j(z)-\x_i(z)\|_2\\
\leq&\langle\nabla f_{t,j}(\x_j(z)),\x_j(z)-\x\rangle-\frac{\alpha}{2}\|\x_j(z)-\x\|_2^2\\
&+G\|\x_j(z)-\bar{\x}(z+1)+\bar{\x}(z+1)-\x_i(z)\|_2\\
\leq&\langle\nabla f_{t,j}(\x_j(z)),\bar{\x}(z+1)-\x\rangle-\frac{\alpha}{2}\|\x_j(z)-\x\|_2^2\\
&+\langle\nabla f_{t,j}(\x_j(z)),\x_j(z)-\bar{\x}(z+1)\rangle+2G\epsilon_z\\
\leq&\langle\nabla f_{t,j}(\x_j(z)),\bar{\x}(z+1)-\x\rangle-\frac{\alpha}{2}\|\x_j(z)-\x\|_2^2\\
&+G\|\x_j(z)-\bar{\x}(z+1)\|_2+2G\epsilon_z\\
\leq&\langle\nabla f_{t,j}(\x_j(z)),\bar{\x}(z+1)-\x\rangle-\frac{\alpha}{2}\|\x_j(z)-\x\|_2^2+3G\epsilon_z
\end{split}
\end{equation}
where the third and last inequalities are due to \eqref{sub_eq8} and \eqref{sub_eq9}. Moreover, for any $\x,\y,\y^\prime$, we have
\begin{equation}
\label{for_ref-D-FTGL2}
\begin{split}
\|\y-\x\|_2^2=&\|\y-\y^\prime\|_2^2+2\langle \y,\y^\prime-\x\rangle+\|\x\|_2^2-\|\y^\prime\|_2^2\\
\geq&2\langle \y,\y^\prime-\x\rangle+\|\x\|_2^2-\|\y^\prime\|_2^2.
\end{split}
\end{equation}
By combining \eqref{for_ref-D-FTGL} with \eqref{for_ref-D-FTGL2}, for any $z\in[T/L]$, $t\in\mathcal{T}_z$, $j\in [n]$, and $\x\in\K$, we have
\begin{equation*}
\begin{split}
&f_{t,j}(\x_i(z))-f_{t,j}(\x)\\
\leq&\langle \nabla f_{t,j}(\x_j(z)),\bar{\x}(z+1)-\x\rangle-\frac{\alpha}{2}\left(2\langle \x_j(z),\bar{\x}(z+1)-\x\rangle+\|\x\|_2^2-\|\bar{\x}(z+1)\|_2^2\right)+3G\epsilon_z\\
=&\langle\nabla f_{t,j}(\x_j(z))-\alpha \x_j(z),\bar{\x}(z+1)-\x\rangle+\frac{\alpha}{2}\left(\|\bar{\x}(z+1)\|_2^2-\|\x\|_2^2\right)+3G\epsilon_z.
\end{split}
\end{equation*}
Finally, from the above inequality and the definition of $\epsilon_z$, it is not hard to verify that
\begin{equation}
\label{for-projection-free-ref}
\begin{split}
&\sum_{z=1}^{T/L}\sum_{t\in\mathcal{T}_z}\sum_{j=1}^nf_{t,j}(\x_i(z))-\sum_{z=1}^{T/L}\sum_{t\in\mathcal{T}_z}\sum_{j=1}^nf_{t,j}(\x)\\
\leq&\sum_{z=1}^{T/L}\sum_{t\in\mathcal{T}_z}\sum_{j=1}^n\left(\langle\nabla f_{t,j}(\x_j(z))-\alpha \x_j(z),\bar{\x}(z+1)-\x\rangle+\frac{\alpha}{2}\left(\|\bar{\x}(z+1)\|_2^2-\|\x\|_2^2\right)+3G\epsilon_z\right)\\
=&n\sum_{z=1}^{T/L}\left(\langle\bar{\d}(z),\bar{\x}(z+1)-\x\rangle+\frac{L\alpha}{2}(\|\bar{\x}(z+1)\|_2^2-\|\x\|_2^2)\right)+3nLG\sum_{z=1}^{T/L}\epsilon_z\\
\overset{\eqref{sub_eq1}}{\leq}&nhR^2+3nLG\left(\sum_{z=2}^{T/L}\frac{3L(G+\alpha R)}{(z-2)L\alpha+2h}+\sum_{z=1}^{T/L}\frac{4L(G+2\alpha R)}{zL\alpha+2h}\right).
\end{split}
\end{equation}

\subsection{Proof of Corollaries \ref{corollary-convex_upper} and \ref{corollary-sconvex_upper}}
By substituting $\alpha=0$ and $h=\sqrt{11LT}G/R$ into \eqref{unified-thm}, we have
\begin{equation*}
\begin{split}
R_{T,i}\leq&3nLG\left(\sum_{z=2}^{T/L}\frac{3LG}{2h}+\sum_{z=1}^{T/L}\frac{2LG}{h}\right)+nhR^2\leq\frac{11nLG^2T}{h}+nhR^2=2nGR\sqrt{11LT}
\end{split}
\end{equation*}
which completes the proof of Corollary \ref{corollary-convex_upper}.

Similarly, by substituting $h=\alpha L$ into \eqref{unified-thm}, we have
\begin{equation*}
\begin{split}
R_{T,i}\leq& 3nLG\left(\sum_{z=2}^{T/L}\frac{3(G+\alpha R)}{z\alpha}+\sum_{z=1}^{T/L}\frac{4(G+2\alpha R)}{(z+2)\alpha}\right)+n\alpha LR^2\\
\leq&\frac{3nLG(7G+11\alpha R)}{\alpha}\sum_{z=1}^{T/L}\frac{1}{z}+n\alpha LR^2\\
\leq&\frac{3nLG(7G+11\alpha R)(1+\ln(T/L))}{\alpha}+n\alpha LR^2
\end{split}
\end{equation*}
which completes the proof of Corollary \ref{corollary-sconvex_upper}.

\subsection{Proof of Theorem \ref{thm_lowerB}}
Recall that \citet{Wan-22-JMLR} have established an $\Omega(nT/\sqrt{C})$ lower bound by extending the classical randomized lower bound for OCO  \citep{Abernethy08} to D-OCO with limited communications. The main limitation of their analysis is that they ignore the topology of the graph $\mathcal{G}$ and the spectral properties of the matrix $P$. To address this limitation, our main idea is to refine their analysis by carefully choosing $\mathcal{G}$ and $P$.

Specifically, let $A\in\mathbb{R}^{n\times n}$ denote the adjacency matrix of $\mathcal{G}$, and let $\delta_i=|N_i|-1$ denote the degree of node $i$. As presented in (8) of \citet{DADO2011}, for any connected undirected graph, there exists a specific choice of the gossip matrix $P$ satisfying Assumption \ref{assum5}, i.e.,
\begin{equation}
\label{choice_of_P}
P=I_n-\frac{1}{\delta_{\max}+1}(D-A)
\end{equation}
where $I_n$ is the identity matrix, $\delta_{\max}=\max\{\delta_1,\dots,\delta_n\}$, and $D=\text{diag}\{\delta_1,\dots,\delta_n\}$. Moreover, \citet{DADO2011} have discussed the connection of the spectral gap $1-\sigma_2(P)$ and the network size $n$ for several classes of interesting networks. Here, we need to utilize the 1-connected cycle graph, i.e., constructing the graph $\mathcal{G}$ by placing the $n$ nodes on a circle and only connecting each node to one neighbor on its right and left. We can derive the following lemma for the 1-connected cycle graph.
\begin{lem}
\label{lem_cycle_graph}
For the 1-connected cycle graph with $n=2(m+1)$ where $m$ denotes a positive integer, the gossip matrix defined in (\ref{choice_of_P}) satisfies
\[\frac{\pi^2}{1-\sigma_2(P)}\leq 4n^2.\]
\end{lem}
Then, we only need to derive a lower bound of $\Omega(n\sqrt{n}T/\sqrt{C})$ since combining it with Lemma \ref{lem_cycle_graph} will complete this proof.
To this end, we set \[f_{t,n-\lceil m/2\rceil+2}(\x)=\cdots=f_{t,n}(\x)=f_{t,1}(\x)=f_{t,2}(\x)=\dots=f_{t,\lceil m/2\rceil}(\x)=0\] and carefully choose other local functions $f_{t,\lceil m/2\rceil+1}(\x),\dots,f_{t,n-\lceil m/2\rceil+1}(\x)$. 

Without loss of generality, we denote the set of communication rounds by $\C=\{c_1,\dots,c_C\}$, where $1\leq c_1<\dots<c_C<T$. According to the topology of the 1-connected cycle graph, it is easy to verify that the learner $1$ cannot receive the information generated by learners $\lceil m/2\rceil+1,\cdots,n-\lceil m/2\rceil+1$ at round $t$ unless there exist $\lceil m/2\rceil$ communication rounds since round $t$. Let $K=\lceil m/2\rceil$, $Z=\lfloor C/K\rfloor$, $c_0=0$, and $c_{(Z+1)K}=T$. The total $T$ rounds can be divided into the following $Z+1$ intervals
\begin{equation}
\label{Communication-division}
[c_0+1,c_{K}],[c_{K}+1,c_{2K}],\dots,[c_{ZK}+1,c_{(Z+1)K}].
\end{equation}
To maximize the impact of the
communication and the topology on the regret of learner $1$, for any $i\in\{0,\dots,Z\}$ and $t\in[c_{iK}+1,c_{(i+1)K}]$, we will set $f_{t,\lceil m/2\rceil+1}(\x)=\dots=f_{t,n-\lceil m/2\rceil+1}(\x)=h_i(\x)$, which implies that the global loss function can be written as
\begin{equation}
\label{lower-v2-eq1}
f_t(\x)=(n-2K+1)h_i(\x).
\end{equation}
Moreover, according to the above discussion, the decisions $\x_1(c_{iK}+1),\dots,\x_1(c_{(i+1)K})$ for any $i\in\{0,\dots,Z\}$ are made before the function $h_i(\x)$ can be revealed to learner $1$. As a result, we can utilize the classical randomized strategy to select $h_i(\x)$ for any $i\in\{0,\dots,Z\}$, and derive an expected lower bound for $R_{T,1}$.

To be precise, we independently select $h_i(\x)=\langle\w_i,\x\rangle$ for any $i\in\{0,\dots,Z\}$, where the coordinates of $\w_i$ are $\pm G/\sqrt{d}$ with probability $1/2$ and $h_i(\x)$ satisfies Assumption \ref{assum4}. It is not hard to verify that
\begin{equation}
\label{lower-v2-eq2}
\begin{split}
&\E_{\w_0,\dots,\w_Z}[R_{T,1}]\\
\overset{\eqref{lower-v2-eq1}}{=}&\E_{\w_0,\dots,\w_Z}\left[\sum_{i=0}^Z\sum_{t=c_{iK}+1}^{c_{(i+1)K}}(n-2K+1)h_i(\x_1(t))-\min\limits_{\x\in\K}\sum_{i=0}^Z\sum_{t=c_{iK}+1}^{c_{(i+1)K}}(n-2K+1)h_i(\x)\right]\\
=&(n-2K+1)\E_{\w_0,\dots,\w_Z}\left[\sum_{i=0}^Z\sum_{t=c_{iK}+1}^{c_{(i+1)K}}\langle\w_i,\x_1(t)\rangle-\min\limits_{\x\in\K}\sum_{i=0}^Z(c_{(i+1)K}-c_{iK})\langle\w_i,\x\rangle\right]\\
=&-(n-2K+1)\E_{\w_0,\dots,\w_Z}\left[\min\limits_{\x\in\K}\sum_{i=0}^Z(c_{(i+1)K}-c_{iK})\langle\w_i,\x\rangle\right]\\
=&-(n-2K+1)\E_{\w_0,\dots,\w_Z}\left[\min\limits_{\x\in\left\{-R/\sqrt{d},R/\sqrt{d}\right\}^d}\left\langle\x,\sum_{i=0}^Z(c_{(i+1)K}-c_{iK})\w_i\right\rangle\right]
\end{split}
\end{equation}
where the third equality is due to $\E_{\w_0,\dots,\w_Z}[\w_i^\top\x_1(t)]=0$ for any $t\in[c_{iK}+1,c_{(i+1)K}]$, and the last equality is because a linear function is minimized at the vertices of the cube.

Then, let $\epsilon_{01},\dots,\epsilon_{0d},\dots,\epsilon_{Z1},\dots,\epsilon_{Zd}$ be independent and identically distributed variables with $\Pr(\epsilon_{ij}=\pm 1)=1/2$ for $i\in\{0,\dots,Z\}$ and $j\in\{1,\dots,d\}$. By combining these notations with \eqref{lower-v2-eq2}, we have
\begin{equation}
\label{eq_lowerB-pre-for-pre}
\begin{split}
\E_{\w_0,\dots,\w_Z}[R_{T,1}]=&-(n-2K+1)\E_{\epsilon_{01},\dots,\epsilon_{Zd}}\left[\sum_{j=1}^d-\frac{R}{\sqrt{d}}\left|\sum_{i=0}^Z(c_{(i+1)K}-c_{iK})\frac{\epsilon_{ij}G}{\sqrt{d}}\right|\right]\\
=&(n-2K+1)RG\E_{\epsilon_{01},\dots,\epsilon_{Z1}}\left[\left|\sum_{i=0}^Z(c_{(i+1)K}-c_{iK})\epsilon_{i1} \right|\right].
\end{split}
\end{equation}
Moreover, by combining \eqref{eq_lowerB-pre-for-pre} with the Khintchine inequality, we have
\begin{equation}
\label{eq_lowerB}
\begin{split}
\E_{\w_0,\dots,\w_Z}[R_{T,1}]\geq&\frac{(n-2K+1)RG}{\sqrt{2}}\sqrt{\sum_{i=0}^Z(c_{(i+1)K}-c_{iK})^2}\\
\geq&\frac{(n-2K+1)RG}{\sqrt{2}}\sqrt{\frac{(c_{(Z+1)K}-c_0)^2}{Z+1}}=\frac{(n-2K+1)RGT}{\sqrt{2(Z+1)}}
\end{split}
\end{equation}
where the second inequality is due to the Cauchy-Schwarz inequality.

Note that the expected lower bound in (\ref{eq_lowerB}) implies that for any D-OCO algorithm with communication rounds $\C=\{c_1,\dots,c_C\}$, there exists a particular choice of $\w_0,\dots,\w_Z$ such that \begin{align*}
R_{T,1}\geq&\frac{(n-2K+1)RGT}{\sqrt{2(Z+1)}}
\geq\frac{n\sqrt{2n}RGT}{4\sqrt{8C+n}}
\end{align*}
where the last inequality is due to
\begin{equation}
\label{share_ieq_1}
\begin{split}
\frac{n-2K+1}{\sqrt{Z+1}}=&\frac{n-2\lceil m/2\rceil+1}{\sqrt{\lfloor C/\lceil m/2\rceil\rfloor+1}}\geq\frac{n-m-1}{\sqrt{ C/\lceil m/2\rceil+1}}=\frac{(m+1)\sqrt{m+1}}{\sqrt{(C/\lceil m/2\rceil+1)(m+1)}}\\
\geq&\frac{(m+1)\sqrt{m+1}}{\sqrt{4C+m+1}}=\frac{n\sqrt{n}}{2\sqrt{8C+n}}.
\end{split}
\end{equation}
If $n\leq 8C+16$, by combining the above result on $R_{T,1}$ with Lemma \ref{lem_cycle_graph}, we have
\[R_{T,1}\geq\frac{n\sqrt{\pi}RGT}{16(1-\sigma_2(P))^{1/4}\sqrt{C+1}}.\]
Otherwise, we have $8C< n-16<n$, and thus $R_{T,1}\geq\frac{nRGT}{4}$.

\subsection{Proof of Lemma \ref{lem_cycle_graph}}
We start this proof by introducing a general lemma regarding the spectral gap of the communication matrix $P$ defined in (\ref{choice_of_P}).
\begin{lem}
\label{lem4_duchi}
(Lemma 4 of \citet{DADO2011}) Let $\delta_i$ denote the degree of each node $i$ in a connected undirected graph $\G$.  For the graph $\G$, the matrix $P$ defined in (\ref{choice_of_P}) satisfies
\[\sigma_2(P)\leq\max\left\{1-\frac{\delta_{\min}}{\delta_{\max}+1}\lambda_{n-1}(\L),\frac{\delta_{\max}}{\delta_{\max}+1}\lambda_{1}(\L)-1\right\}\]
where $\delta_{\min}=\min\{\delta_1,\dots,\delta_n\}$, $\delta_{\max}=\max\{\delta_1,\dots,\delta_n\}$, $\L$ denotes the normalized graph Laplacian of $\mathcal{G}$, and $\lambda_i(\L)$ denotes the $i$-th largest real eigenvalue of $\L$.
\end{lem}
Moreover, \citet{DADO2011} have proved that $\L$ has the following eigenvalues
\[\left\{1-\left.\cos\left(\frac{2\pi i}{n}\right)\right|i=1,\dots,n\right\}\]
for the 1-connected cycle graph.

Therefore, because of $n=2(m+1)$, it is easy to verify that
\[\lambda_1(\L)=1-\cos\left(\frac{2(m+1)\pi}{n}\right)=1-\cos(\pi)=2.\]
Then, because of $n=2(m+1)$ and $\cos(x)=\cos(2\pi-x)$ for any $x$, we have
\[\lambda_{n-1}(\L)=\min\left\{1-\cos\left(\frac{2\pi}{n}\right),1-\cos\left(\frac{2\pi(n-1)}{n}\right)\right\}=1-\cos\left(\frac{\pi}{m+1}\right)\geq\frac{\pi^2}{4(m+1)^2}.\]
Since the 1-connected cycle graph also satisfies that $\delta_{\max}=\delta_{\min}=2$, by using Lemma \ref{lem4_duchi}, we have
\[\sigma_2(P)\leq\max\left\{1-\frac{2}{3}\lambda_{n-1}(\L),\frac{1}{3}\right\}=1-\frac{2}{3}\lambda_{n-1}(\L)\leq1-\frac{\pi^2}{6(m+1)^2}\]
where the equality is due to $\lambda_{n-1}(\L)\leq1-\cos(\pi/2)\leq 1$. 

Finally, it is easy to verify that
\[\frac{\pi^2}{1-\sigma_2(P)}\leq 6(m+1)^2\leq 4n^2.\]

\subsection{Proof of Theorem \ref{thm_lowerB-sc}}
The proof of Theorem \ref{thm_lowerB-sc} is similar to the proof of Theorem \ref{thm_lowerB}. The main modification is to make the previous local functions $\alpha$-strongly convex by adding a term $\frac{\alpha}{2}\|\x\|_2^2$. 

To be precise, let $K=\lceil m/2\rceil$, $Z=\lfloor C/K\rfloor$, $c_0=0$, and $c_{(Z+1)K}=T$. We still denote the set of communication rounds by $\C=\{c_1,\dots,c_C\}$ where $1\leq c_1<\dots<c_C<T$, and divide the total $T$ rounds into $Z+1$ intervals defined in \eqref{Communication-division}. At each round $t$, we simply set
\[f_{t,n-\lceil m/2\rceil+2}(\x)=\cdots=f_{t,n}(\x)=f_{t,1}(\x)=f_{t,2}(\x)=\cdots=f_{t,\lceil m/2\rceil}(\x)=\frac{\alpha}{2}\|\x\|_2^2\] which satisfies Assumption \ref{assum4} with $G=2\alpha R$ and the definition of $\alpha$-strongly convex functions.
Moreover, for any $i\in\{0,\dots,Z\}$ and $t\in[c_{iK}+1,c_{(i+1)K}]$, we set \[f_{t,\lceil m/2\rceil+1}(\x)=\dots=f_{t,n-\lceil m/2\rceil+1}(\x)=h_i(\x)=\langle \w_i,\x\rangle+\frac{\alpha}{2}\|\x\|^2_2\]
where the coordinates of $\w_i$ are $\pm \alpha R/\sqrt{d}$ with probability $1/2$. Note that $h_i(\x)$ satisfies Assumption \ref{assum4} with $G=2\alpha R$ and Assumption \ref{scvx-assum}.
Following the proof of Theorem \ref{thm_lowerB}, we set $\G$ as the $1$-connected cycle graph, which ensures that the decisions $\x_1(c_{iK}+1),\dots,\x_1(c_{(i+1)K})$ are independent of $\w_i$.

Then, let $\bar{\w}=\frac{1}{\alpha T}\sum_{i=0}^Z(c_{(i+1)K}-c_{iK})\w_i$. The total loss for any $\x\in\K$ equals to
\begin{equation}
\label{toal_loss-sc}
\begin{split}
\sum_{t=1}^Tf_t(\x)
=&\sum_{i=0}^Z(c_{(i+1)K}-c_{iK})\left((n-2K+1)\langle \w_i,\x\rangle+\frac{\alpha n}{2}\|\x\|_2^2\right)\\
=&\alpha(n-2K+1)T\langle \bar{\w},\x\rangle+\frac{\alpha nT}{2}\|\x\|_2^2\\
=&\frac{\alpha T}{2}\left(\left\|\sqrt{n}\x+\frac{(n-2K+1)}{\sqrt{n}}\bar{\w}\right\|_2^2-\left\|\frac{(n-2K+1)}{\sqrt{n}}\bar{\w}\right\|_2^2\right).
\end{split}
\end{equation}
According to the definition of $\w_i$, the absolute value of each element in $-\frac{n-2K+1}{n}\bar{\w}$ is bounded by
\begin{align*}
\frac{n-2K+1}{n\alpha T}\sum_{i=0}^Z\frac{(c_{(i+1)K}-c_{iK})\alpha R}{\sqrt{d}}=\frac{(n-2K+1)R}{n\sqrt{d}}\leq\frac{R}{\sqrt{d}}
\end{align*}
which implies that $-\frac{n-2K+1}{n}\bar{\w}$ belongs to $\K=[-R/\sqrt{d},R/\sqrt{d}]^d$.

By further combining with (\ref{toal_loss-sc}), we have
\begin{equation*}
\argmin_{\x\in\K}\sum_{t=1}^Tf_t(\x)=-\frac{n-2K+1}{n}\bar{\w} \text{ and }\min_{\x\in\K}\sum_{t=1}^Tf_t(\x)=-\frac{\alpha T}{2}\left\|\frac{(n-2K+1)}{\sqrt{n}}\bar{\w}\right\|_2^2.
\end{equation*}
As a result, it is not hard to verify that
\begin{equation*}
\begin{split}
&\E_{\w_0,\dots,\w_Z}\left[R_{T,1}\right]\\
=&\E_{\w_0,\dots,\w_Z}\left[\sum_{i=0}^Z\sum_{t=c_{iK}+1}^{c_{(i+1)K}}\left((n-2K+1)\langle\w_i,\x_1(t)\rangle+\frac{\alpha n}{2}\|\x_1(t)\|_2^2\right)+\frac{\alpha T}{2}\left\|\frac{(n-2K+1)}{\sqrt{n}}\bar{\w}\right\|_2^2\right]\\
\geq&\E_{\w_0,\dots,\w_Z}\left[\sum_{i=0}^Z\sum_{t=c_{iK}+1}^{c_{(i+1)K}}(n-2K+1)\langle\w_i,\x_1(t)\rangle+\frac{\alpha(n-2K+1)^2T}{2n}\|\bar{\w}\|_2^2\right]\\
=&\E_{\w_0,\dots,\w_Z}\left[\frac{\alpha(n-2K+1)^2T}{2n}\|\bar{\w}\|_2^2\right]
\end{split}
\end{equation*}
where  the last equality is due to $\E_{\w_0,\dots,\w_Z}[\w_i^\top\x_1(t)]=0$ for any $t\in[c_{iK}+1,c_{(i+1)K}]$.

Next, let $\epsilon_{01},\dots,\epsilon_{0d},\dots,\epsilon_{Z1},\dots,\epsilon_{Zd}$ be independent and identically distributed variables with $\Pr(\epsilon_{ij}=\pm 1)=1/2$ for $i\in\{0,\dots,Z\}$ and $j\in\{1,\dots,d\}$. By combining the definition of $\bar{\w}$ with the above inequality, we further have
\begin{equation}
\label{eq_lowerB-sc}
\begin{split}
\E_{\w_0,\dots,\w_Z}\left[R_{T,1}\right]
\geq&\frac{(n-2K+1)^2}{2\alpha nT}\E_{\w_0,\dots,\w_Z}\left[\left\|\sum_{i=0}^Z(c_{(i+1)K}-c_{iK})\w_i\right\|_2^2\right]\\
=&\frac{(n-2K+1)^2}{2\alpha nT}\E_{\epsilon_{01},\dots,\epsilon_{Zd}}\left[\sum_{j=1}^d\left|\sum_{i=0}^Z(c_{(i+1)K}-c_{iK})\frac{\epsilon_{ij}\alpha R}{\sqrt{d}}\right|^2\right]\\
=&\frac{\alpha(n-2K+1)^2R^2}{2 nT}\E_{\epsilon_{01},\dots,\epsilon_{Z1}}\left[\left|\sum_{i=0}^Z(c_{(i+1)K}-c_{iK})\epsilon_{i1} \right|^2\right]\\
=&\frac{\alpha(n-2K+1)^2R^2}{2 nT}\sum_{i=0}^Z(c_{(i+1)K}-c_{iK})^2\geq \frac{\alpha(n-2K+1)^2R^2T}{2n(Z+1)}
\end{split}
\end{equation}
where the inequality is due to the Cauchy-Schwarz inequality and $(c_{(Z+1)K}-c_0)^2=T^2$.

The expected lower bound in (\ref{eq_lowerB-sc}) implies that for any D-OCO algorithm with communication rounds $\C=\{c_1,\dots,c_C\}$, there exists a particular choice of $\w_0,\dots,\w_Z$ such that \begin{align*}
R_{T,1}\geq\frac{\alpha(n-2K+1)^2R^2T}{2n(Z+1)}\overset{\eqref{share_ieq_1}}{\geq}\frac{\alpha n^2R^2T}{8(8C+n)}.
\end{align*}
If $n\leq 8C+16$, according to Lemma \ref{lem_cycle_graph}, by using the gossip matrix $P$ defined in (\ref{choice_of_P}), we have
\[R_{T,1}\geq\frac{\alpha \pi n R^2T}{256(C+1)\sqrt{1-\sigma_2(P)}}.\]
Otherwise, we have $8C<n-16<n$, and thus $R_{T,1}\geq\frac{\alpha nR^2T}{16}$.

\subsection{Proof of Theorem \ref{imp-thm_lowerB-sc}}
\label{sec-Proof-Theorem4}
Compared with the proof of Theorem \ref{thm_lowerB-sc}, the main difference of this proof is to focus on the case with $C=T-1$ and redefine the loss functions as well as the decision set. Specifically, for any D-OCO algorithm, we still denote the sequence of decisions made by the local learner $1$ as $\x_1(1),\dots,\x_1(T)$, but divide the total $T$ rounds into the following $Z+1$ intervals
\begin{equation}
\label{no-communication-intervals}
[c_0+1,c_1],[c_1+1,c_2],\dots,[c_{Z}+1,c_{Z+1}]
\end{equation}
where $Z=\lfloor (T-1)/K\rfloor$, $K=\lceil m/2\rceil$, $c_{Z+1}=T$, and $c_i=iK$ for $i=0,\dots,Z$. At each round $t$, we first simply set
\begin{equation*}
f_{t,n-\lceil m/2\rceil+2}(\x)=\dots=f_{t,n}(\x)=f_{t,1}(\x)=f_{t,2}(\x)=\dots=f_{t,\lceil m/2\rceil}(\x)=\frac{\alpha}{2}\|\x\|_2^2
\end{equation*}
which is $\alpha$-strongly convex and satisfies Assumption \ref{assum4} with $G=\alpha R$ over the set $\K=[0,R/\sqrt{d}]^d$. Then, let $\mathcal{B}_p$ denote the Bernoulli distribution with probability of obtaining $1$ equal to $p$. For any $i\in\{0,\dots,Z\}$ and $t\in[c_{i}+1,c_{i+1}]$, we set
\begin{equation*}
f_{t,\lceil m/2\rceil+1}(\x)=\dots=f_{t,n-\lceil m/2\rceil+1}(\x)=h_i(\x)=\frac{\alpha}{2}\left\|\x-\frac{R\w_i}{\sqrt{d}}\right\|^2_2
\end{equation*}
where $\w_i$ is sampled from the vector set $\{\mathbf{0},\mathbf{1}\}$ according to $\mathcal{B}_p$, i.e., $\Pr(\w_i=\mathbf{1})=p$. It is easy to verify that $h_i(\x)$ also satisfies the definition of $\alpha$-strongly convex functions and Assumption \ref{assum4} with $G=\alpha R$ over the set $\K=[0,R/\sqrt{d}]^d$. Then, for any $i\in\{0,\dots,Z\}$ and $t\in[c_{i}+1,c_{i+1}]$, the global loss function in each round can be written as
\begin{equation*}
\begin{split}
f_t(\x)=&\sum_{j=1}^nf_{t,j}(\x)=\frac{\alpha(n-2K+1)}{2}\left\|\x-\frac{R\w_i}{\sqrt{d}}\right\|_2^2+\frac{\alpha(2K-1)}{2}\|\x\|_2^2\\
=&\frac{\alpha n}{2}\left\|\x\right\|_2^2-\frac{\alpha(n-2K+1)R}{\sqrt{d}}\left\langle\x,\w_i\right\rangle+\frac{\alpha(n-2K+1)R^2}{2d}\left\|\w_i\right\|_2^2
\end{split}
\end{equation*}
whose expectation is
\begin{equation*}
\begin{split}
\E_{\w_i}\left[f_t(\x)\right]
=&\frac{\alpha n}{2}\left\|\x\right\|_2^2+\frac{\alpha(n-2K+1)R}{\sqrt{d}}\left\langle\x,\mathbf{p}\right\rangle+\frac{\alpha(n-2K+1)R^2}{2d}\left\langle\mathbf{1},\mathbf{p}\right\rangle\\
=&\frac{\alpha n}{2}\left\|\x-\frac{(n-2K+1)R\mathbf{p}}{n\sqrt{d}}\right\|_2^2+\frac{\alpha(n-2K+1)R^2}{2d}\left\langle\mathbf{1}-\frac{(n-2K+1)\mathbf{p}}{n},\mathbf{p}\right\rangle
\end{split}
\end{equation*}
where $\mathbf{p}$ denotes the vector with the same value of $p$ in each dimension. For brevity, let $F(\x)=\E_{\w_i}\left[f_t(\x)\right]$. Because of $p\in[0,1]$ and $\K=[0,R/\sqrt{d}]^d$, this function can be simply minimized by
\begin{equation*}
\x^\ast=\frac{(n-2K+1)R\mathbf{p}}{n\sqrt{d}}\in\K.
\end{equation*}
which implies that any $\x\in\K$ has
\begin{equation}
\label{extended-eq1}
F(\x)-F(\x^\ast)=\frac{\alpha n}{2}\left\|\x-\frac{(n-2K+1)R\mathbf{p}}{n\sqrt{d}}\right\|_2^2\geq 0.
\end{equation}
Moreover, it is not hard to verify that
\begin{equation}
\label{extended-eq2}
\begin{split}
\E_{\w_0,\dots,\w_Z}\left[\min_{\x\in\K}\sum_{i=0}^{Z}\sum_{t=c_i+1}^{c_{i+1}}f_t(\x)\right]\leq\E_{\w_0,\dots,\w_Z}\left[\sum_{i=0}^{Z}\sum_{t=c_i+1}^{c_{i+1}}f_t(\x^\ast)\right]=\sum_{i=0}^{Z}\sum_{t=c_i+1}^{c_{i+1}}F(\x^\ast).
\end{split}
\end{equation}
Following the proof of Theorem \ref{thm_lowerB-sc}, we continue to set $\G$ as the $1$-connected cycle graph, which ensures that the decisions $\x_1(c_{i}+1),\dots,\x_1(c_{i+1})$ are independent of $\w_i$. Therefore, we have
\begin{equation}
\label{extended-eq3}
\begin{split}
\E_{\w_0,\dots,\w_Z}[R_{T,1}]=&\E_{\w_0,\dots,\w_Z}\left[\sum_{i=0}^{Z}\sum_{t=c_i+1}^{c_{i+1}}f_t(\x_1(t))-\min_{\x\in\K}\sum_{i=0}^{Z}\sum_{t=c_i+1}^{c_{i+1}}f_t(\x)\right]\\
=&\E_{\w_0,\dots,\w_Z}\left[\sum_{i=0}^{Z}\sum_{t=c_i+1}^{c_{i+1}}F(\x_1(t))\right]-\E_{\w_0,\dots,\w_Z}\left[\min_{\x\in\K}\sum_{i=0}^{Z}\sum_{t=c_i+1}^{c_{i+1}}f_t(\x)\right]\\
\overset{\eqref{extended-eq2}}{\geq}&\E_{\w_0,\dots,\w_Z}\left[\sum_{i=0}^{Z}\sum_{t=c_i+1}^{c_{i+1}}F(\x_1(t))-\sum_{i=0}^{Z}\sum_{t=c_i+1}^{c_{i+1}}F(\x^\ast)\right].
\end{split}
\end{equation}
To lower bound the right side of \eqref{extended-eq3}, we assume that the D-OCO algorithm is deterministic without loss of generality.\footnote{As in the lower bound analysis of \citet{Epoch-GD}, even if the algorithm is randomized, it can be viewed as a deterministic one by fixing its random seed.} Note that all local functions $\{f_{t,1}(\x),\dots,\{f_{t,n}(\x)\}$ for any $t\in[c_{i}+1,c_{i+1}]$ are either deterministic or parameterized by the same random vector $\w_i$ drawn according to $\mathcal{B}_p$. Therefore, for any $t\in[c_i+1,c_{i+1}]$, $\x_1(t)$ actually can be specified by a bit string $X\in\{0,1\}^{i}$ drawn from $\mathcal{B}_p^i$, i.e., the product measure on $\{0,1\}^i$ induced by taking $i$ independent trials from $\mathcal{B}_p$. To be precise, the local learner $1$ of the D-OCO algorithm at any round $t\in[c_{i}+1,c_{i+1}]$ can be denoted as a mapping function $\mathcal{A}_t(\cdot):\{0,1\}^{i}\mapsto\K$ such that $\x_1(t)=\mathcal{A}_t(X)$.
Moreover, one should note that the value $p$ in the above procedures can be replaced by another value $p^\prime$, and the corresponding random vectors can be rewritten as $\w^\prime_0,\dots,\w^\prime_Z$ and $\x_1^\prime(1),\dots,\x_1^\prime(T)$. Similarly, $\x_1^\prime(t)$ for any round $t\in[c_{i}+1,c_{i+1}]$ can be specified by a bit string $X^\prime\in\{0,1\}^{i}$ drawn from $\mathcal{B}_{p^\prime}^i$, i.e., $\x_1^\prime(t)=\mathcal{A}_t(X^\prime)$. Interestingly, following \citet{Epoch-GD}, we can show that the expected instantaneous regret of the local learner $1$ on at least one of the two distributions parameterized by appropriate $p$ and $p^\prime$ must be large.


\begin{lem}
\label{extended-lem1}
Fix a block $i$ and let $\epsilon\leq\frac{1}{32\sqrt{i+1}}$ be a parameter. Assume that $p,p^\prime\in\left[\frac{1}{4},\frac{3}{4}\right]$ such that $2\epsilon\leq|p-p^\prime|\leq 4\epsilon$. Following the above notations, for any $t\in[c_i+1,c_{i+1}]$, we have
\begin{equation*}
\begin{split}
\E_{X}\left[\left\|\mathcal{A}_t(X)-\xi\mathbf{p}\right\|_2^2\right]+\E_{X^\prime}\left[\left\|\mathcal{A}_t(X^\prime)-\xi\mathbf{p}^\prime\right\|_2^2\right]\geq \frac{d(\xi\epsilon)^2}{4}
\end{split}
\end{equation*}
where $\xi=(n-2K+1)R/(n\sqrt{d})$ and $\mathbf{p}^\prime$ denotes the vector with the same value of $p^\prime$ in each dimension.
\end{lem}
Let $M=\left\lfloor\log_{16}(15Z+16)-1\right\rfloor$, and it is not hard to verify that $M\geq 1$ due to $16n+1\leq T$. To exploit the above lemma, we further divide the first $Z^\prime=\frac{1}{15}(16^{M+1}-16)$ intervals into $M$ epochs with the length $16,16^2,\dots,16^M$. More specifically, the $m$-th epoch $E_{m}$ consists of the intervals $\frac{1}{15}(16^m-16),\dots,\frac{1}{15}(16^{m+1}-16)-1$. Then, for these $M$ epochs, we can prove the following lemma based on Lemma \ref{extended-lem1}.
\begin{lem}
\label{extended-lem2}
Following the notations used in Lemma \ref{extended-lem1}, there exists a collection of nested intervals, $\left[\frac{1}{4},\frac{3}{4}\right]\supseteq I_1 \supseteq I_2\supseteq \dots \supseteq I_M$, such that interval $I_m$ corresponds to epoch $m$, with the property that $I_m$ has length $4^{-(m+3)}$, and for every $p\in I_m$, we have \[\E_{X}\left[\left\|\mathcal{A}_t(X)-\xi\mathbf{p}\right\|_2^2\right]\geq \frac{16^{-(m+3)}d\xi^2}{8}\] over at least half the rounds $t$ in intervals of epoch $m$.
\end{lem}
From Lemma \ref{extended-lem2}, there exists a value of $p\in\cap_{m\in[M]}I_m$ such that
\begin{equation}
\label{extended-eq13}
\begin{split}
\E_{\w_0,\dots,\w_Z}[R_{T,1}]
\geq&\E_{\w_0,\dots,\w_Z}\left[\sum_{i=0}^{Z}\sum_{t=c_i+1}^{c_{i+1}}\frac{\alpha n}{2}\left\|\x_1(t)-\frac{(n-2K+1)R\mathbf{p}}{n\sqrt{d}}\right\|_2^2\right]\\
\geq&\E_{\w_0,\dots,\w_Z}\left[\sum_{i=0}^{Z^\prime}\sum_{t=c_i+1}^{c_{i+1}}\frac{\alpha n}{2}\left\|\x_1(t)-\frac{(n-2K+1)R\mathbf{p}}{n\sqrt{d}}\right\|_2^2\right]\\
=&\E_{\w_0,\dots,\w_Z}\left[\sum_{m=1}^{M}\sum_{i\in E_m}\sum_{t=c_i+1}^{c_{i+1}}\frac{\alpha n}{2}\left\|\x_1(t)-\frac{(n-2K+1)R\mathbf{p}}{n\sqrt{d}}\right\|_2^2\right]\\
=&\sum_{m=1}^{M}\sum_{i\in E_m}\sum_{t=c_i+1}^{c_{i+1}}\E_{X}\left[\frac{\alpha n}{2}\left\|\mathcal{A}_t(X)-\frac{(n-2K+1)R\mathbf{p}}{n\sqrt{d}}\right\|_2^2\right]\\
\geq &\sum_{m=1}^{M}\frac{\left(c_{\frac{1}{15}(16^{m+1}-16)}-c_{\frac{1}{15}(16^m-16)}\right)16^{-(m+3)}\alpha(n-2K+1)^2R^2}{32n}\\
= &\sum_{m=1}^{M}\frac{16^{-4}\alpha K(n-2K+1)^2R^2}{2n}=\frac{16^{-4}\alpha M K(n-2K+1)^2R^2}{2n}
\end{split}
\end{equation}
where the first inequality is due to \eqref{extended-eq1} and \eqref{extended-eq3}, and the third equality is due to $c_i=iK$ for $i\leq Z$. Moreover, because of the definitions of $M,Z,K$, we have
\begin{equation}
\label{extended-eq14}
\begin{split}
\frac{M K(n-2K+1)^2}{2n}\geq &\frac{(\log_{16}(15Z+16)-2)m(n-m-1)^2}{4n}\\
=&\frac{(\log_{16}(15\lfloor (T-1)/K\rfloor+16)-2)(n-2)n}{32}\\
\geq&\frac{(\log_{16}(30(T-1)/n)-2)(n-2)n}{32}.
\end{split}
\end{equation}
By combining \eqref{extended-eq13} and \eqref{extended-eq14} with Lemma \ref{lem_cycle_graph}, there exists a particular choice of $\w_0,\dots,\w_Z$ such that
\begin{equation}
\label{for_dis_low}
\begin{split}
R_{T,1}
\geq &\frac{16^{-5}\alpha\pi (\log_{16}(30(T-1)/n)-2)(n-2)R^2}{4\sqrt{1-\sigma_2(P)}}
\end{split}
\end{equation}
which completes the proof. 

Additionally, we notice that it is also appealing to extend the lower bound in \eqref{for_dis_low} into the setting with only $C$ communication rounds. Following the proof of Theorem \ref{thm_lowerB-sc}, a natural idea is to divide the total $T$ rounds into $Z+1$ intervals defined in \eqref{Communication-division}, rather than \eqref{no-communication-intervals}, and then repeat the proof steps outlined above. However, we want to emphasize that the last two equalities in \eqref{extended-eq13} require the number of rounds in the first $Z$ intervals to be the same, which is not necessarily satisfied by \eqref{Communication-division}. This is because the algorithm can allocate their $C$ communication rounds arbitrarily. Therefore, instead of simply using this natural idea, new analytical tools may be required for the extension, which will be investigated in the future.  
  
\subsection{Proof of Lemma \ref{extended-lem1}}
This proof is heavily inspired by the proof of Lemma 10 in \citet{Epoch-GD}, but requires specific modifications to generalize their result from one dimension to high dimensions. Specifically, we start this proof by supposing that there exists an integer $j$ such that
\begin{equation}
\label{extended-eq5}
\begin{split}
\E_{X}\left[\left(\mathcal{A}_{t,j}(X)-\xi p\right)^2\right]+\E_{X^\prime}\left[\left(\mathcal{A}_{t,j}(X^\prime)-\xi p^\prime\right)^2\right]< \frac{(\xi\epsilon)^2}{4}
\end{split}
\end{equation}
where $\mathcal{A}_{t,j}(X)$ denotes the $j$-th element in the vector $\mathcal{A}_{t}(X)$. 

Moreover, we first consider the case with $p^\prime\geq p+2\epsilon$. Let $\Pr_p[\cdot]$ and $\Pr_{p^\prime}[\cdot]$ denote the probability of an event under the distribution $\mathcal{B}_p^i$ and $\mathcal{B}_{p^\prime}^i$, respectively. By combining \eqref{extended-eq5} with Markov's inequality, we have
\begin{equation*}
\Pr_p\left[\left(\mathcal{A}_{t,j}(X)-\xi p\right)^2<(\xi\epsilon)^2\right]\geq 3/4
\end{equation*}
which implies that
\begin{equation}
\label{extended-eq6}
\Pr_p\left[\mathcal{A}_{t,j}(X)<\xi(p+\epsilon)\right]\geq 3/4.
\end{equation}
Similarly, we can show that
\begin{equation}
\label{extended-eq7}
\Pr_{p^\prime}\left[\mathcal{A}_{t,j}(X^\prime)>\xi(p+\epsilon)\right]\geq\Pr_{p^\prime}\left[\mathcal{A}_{t,j}(X^\prime)>\xi(p^\prime-\epsilon)\right]\geq 3/4
\end{equation}
where the first inequality is due to $p^\prime\geq p+2\epsilon$. 

Now, we define an event
\[
\mathcal{E}=\left\{Y\in\{0,1\}^{i}:A_{t,j}(Y)>\xi(p+\epsilon)\right\}.
\]
It is easy to verify that
\begin{equation}
\label{extended-eq8}
\left|\Pr\limits_p[\mathcal{E}]-\Pr\limits_{p^\prime}[\mathcal{E}]\right|=\Pr\limits_{p^\prime}[\mathcal{E}]-\Pr\limits_p[\mathcal{E}]>\frac{1}{2}
\end{equation}
where the lase inequality is due to $\Pr_p[\mathcal{E}]< 1/4$ derived from \eqref{extended-eq6} and $\Pr_{p^\prime}[\mathcal{E}]\geq 3/4$ derived from \eqref{extended-eq7}. 

Then, let $d_{\text{TV}}(\mathcal{B}_p^i,\mathcal{B}_{p^\prime}^i)$ denote the total variation distance between the two distributions $\mathcal{B}_p^i$ and $\mathcal{B}_{p^\prime}^i$ on the same probability space, i.e., \[d_{\text{TV}}(\mathcal{B}_p^i,\mathcal{B}_{p^\prime}^i)=\sup_{\mathcal{E}}\left|\Pr\limits_p[\mathcal{E}]-\Pr\limits_{p^\prime}[\mathcal{E}]\right|.\]
Note that \citet{Epoch-GD} have provided an upper bound on $d_{\text{TV}}(\mathcal{B}_p^i,\mathcal{B}_{p^\prime}^i)$.
\begin{lem}
\label{9-hazan}
(Lemma 9 of \citet{Epoch-GD}) Let $p,p^\prime\in\left[\frac{1}{4},\frac{3}{4}\right]$ such that $|p^\prime-p|\leq 1/8$. Then, it holds that
\[d_{\text{TV}}(\mathcal{B}_p^i,\mathcal{B}_{p^\prime}^i)\leq 2\sqrt{(p^\prime-p)^2i}\]
for any integer $i>0$.
\end{lem}
Recall that $|p-p^\prime|\leq 4\epsilon\leq 1/8$. Then, from Lemma \ref{9-hazan}, we have
\begin{equation*}
\left|\Pr\limits_p[\mathcal{E}]-\Pr\limits_{p^\prime}[\mathcal{E}]\right|\leq d_{\text{TV}}(\mathcal{B}_p^i,\mathcal{B}_{p^\prime}^i)\leq 2\sqrt{(p^\prime-p)^2i}\leq 2\sqrt{16\epsilon^2i}\leq\frac{1}{4}
\end{equation*}
which contradicts \eqref{extended-eq8}, and thus implies that our assumption about \eqref{extended-eq5} is unavailable. Additionally, it is easy to construct the same contradiction in the case with $p\geq p^{\prime}+2\epsilon$. Therefore, we have
\begin{equation}
\label{extended-eq9}
\begin{split}
\E_{X}\left[\left(\mathcal{A}_{t,j}(X)-\xi p\right)^2\right]+\E_{X^\prime}\left[\left(\mathcal{A}_{t,j}(X^\prime)-\xi p^\prime\right)^2\right]\geq \frac{(\xi\epsilon)^2}{4}
\end{split}
\end{equation}
for any $j\in[d]$.
Finally, by summing both sides of \eqref{extended-eq9} over $j\in[d]$, we complete this proof.

\subsection{Proof of Lemma \ref{extended-lem2}}
This lemma can be proved by slightly modifying the proof of Lemma 11 in \citet{Epoch-GD}. Here, we include the detailed proof for the completeness. Following \citet{Epoch-GD}, we will iteratively build the required interval $I_m$ for $m=1,\dots,M$. Specifically, we first select an arbitrary interval $I_0=\left[a,a+4^{-4}\right]$ of length $4^{-3}$ inside $\left[\frac{1}{4},\frac{3}{4}\right]$. To find the required interval for $m=1$, we divide $I_{m-1}$ into four equal quarters of length $4^{-{(m+3)}}$, and show that either the first quarter $Q_1=\left[a,a+4^{-(m+3)}\right]$ or the last quarter $Q_4=\left[a+3\cdot4^{-(m+3)},a+4\cdot4^{-(m+3)}\right]$ is a valid choice for $I_m$.

To this end, we suppose that $Q_1$ is not a valid choice for $I_m$, i.e., there exist some $p\in Q_1$ such that it holds \[\E_{X}\left[\left\|\mathcal{A}_{t}(X)-\xi\mathbf{p}\right\|_2^2\right]< \frac{16^{-(m+3)}d\xi^2}{8}\]
for more than half the rounds $t$ in intervals of epoch $m$. Then, we define the following set
\begin{equation}
\label{extended-eq10}
H=\bigcup_{i\in E_m}\left\{t\in[c_{i}+1,c_{i+1}]:\E_{X}\left[\left\|\mathcal{A}_{t}(X)-\xi\mathbf{p}\right\|_2^2\right]< \frac{16^{-(m+3)}d\xi^2}{8}\right\}
\end{equation}
where $E_m=\left\{\frac{1}{15}(16^m-16),\dots,\frac{1}{15}(16^{m+1}-16)-1\right\}$. In the following, we proceed to prove that for all $p^\prime\in Q_4$ and $t\in H$, the following inequality must hold
\begin{equation}
\label{extended-eq11}
\E_{X^\prime}\left[\left\|\mathcal{A}_{t}(X^\prime)-\xi\mathbf{p}^\prime\right\|_2^2\right]\geq \frac{16^{-(m+3)}d\xi^2}{8}
\end{equation}
which implies that $Q_4$ is a valid choice for $I_m$ due to $|H|$ is larger than half the rounds in intervals of epoch $m$.

To be precise, we fix any $p^\prime\in Q_4$ and $t\in H$, where $t$ must belong to $[c_{i}+1,c_{i+1}]$ for some $i\in E_m$. By defining $\epsilon=4^{-(m+3)}$, it is easy to verify that $4(i+1)\leq 16^{m+1}$ due to the definition of $E_m$, and we thus have $\epsilon\leq 1/(32\sqrt{i+1})$. Additionally, we have $2\epsilon \leq |p-p^\prime|\leq 4\epsilon$ due to $p\in Q_1$ and $p^\prime \in Q_4$. Therefore, from Lemma \ref{extended-lem1}, we have
\begin{equation}
\label{extended-eq12}
\begin{split}
\E_{X}\left[\left\|\mathcal{A}_t(X)-\xi\mathbf{p}\right\|_2^2\right]+\E_{X^\prime}\left[\left\|\mathcal{A}_t(X^\prime)-\xi\mathbf{p}^\prime\right\|_2^2\right]\geq \frac{16^{-(m+3)}d\xi^2}{4}.
\end{split}
\end{equation}
Then, the previously mentioned \eqref{extended-eq11} can be simply derived by combining \eqref{extended-eq10} and \eqref{extended-eq12}. Finally, we note that for any $m=2,\dots,M$, the required interval $I_m$ can be built one by one by starting the division from the valid $I_{m-1}$ and repeating the above procedures.

\subsection{Proof of Theorem \ref{upper-projection-free}}
Following the definitions of $\bar{\x}(z)$, $\bar{\z}(z)$, $\bar{\d}(z)$, and $\ell_z(\x)$ in the proof of Theorem \ref{sconvex_upper}, we only need to analyze the distance $\|\x_i(z)-\bar{\x}(z+1)\|_2$. To this end, we first define
\begin{equation*}
\x_i^\ast(z)=\argmin_{\x\in\K}\left\langle\x, \z_i(z-1)\right\rangle+\frac{(z-2)L\alpha}{2}\|\x\|_2^2+h\|\x\|_2^2.
\end{equation*}
for any $z=2,\dots,T/L$, which is exactly the same as $\x_i(z)$ generated by Algorithm \ref{ADOA}. Note that the distance $\|\x_i^\ast(z)-\bar{\x}(z+1)\|_2$ has been analyzed in the proof of Theorem \ref{sconvex_upper}. However, due to the use of CG, $\x_i(z)$ generated by Algorithm \ref{Projection-Free} is only an approximation of $\x_i^\ast(z)$. Moreover, according to Algorithm \ref{Projection-Free}, now $\x_i(z)$ is computed based on $\z_{i}(z-2)$, rather than $\z_i(z-1)$. As a result, we first upper bound $\|\x_i(z)-\bar{\x}(z+1)\|_2$ as 
\begin{equation}
\label{Projection-Free-eq1}
\begin{split}
\|\x_i(z)-\bar{\x}(z+1)\|_2\leq&\|\x_i(z)-\x_i^\ast(z-1)\|_2+\|\x_i^\ast(z-1)-\bar{\x}(z+1)\|_2\\
\leq&\|\x_i(z)-\x_i^\ast(z-1)\|_2+\|\x_i^\ast(z-1)-\bar{\x}(z)\|_2\\
&+\|\bar{\x}(z)-\bar{\x}(z+1)\|_2
\end{split}
\end{equation}
for any $z=3,\dots,T/L$. 

To bound the first term in the right side of \eqref{Projection-Free-eq1}, we introduce the following lemma regarding the convergence property of CG.
\begin{lem}
\label{lem-CG}
(Derived from Theorem 1 of \citet{Revist_FW}) If $F(\x):\K\mapsto\mathbb{R}$ is a convex and $\beta$-smooth function, and $\|\x\|_2\leq R$ for any $\x\in\K$, Algorithm \ref{CG} with $K\geq 1$ ensures
\[
F(\x_{\oo})-F(\x^\ast)\leq \frac{8\beta R^2}{L+2}.
\]
where $\x^\ast\in\argmin_{\x\in\K}F(x)$.
\end{lem}
It is easy to verify that $F_{z-1,i}(\x)$ defined in Algorithm \ref{Projection-Free} is $((z-3)L\alpha+2h)$-strongly convex and $((z-3)L\alpha+2h)$-smooth over $\K$, for any $z=3,\dots,T/L$. Moreover, due to \eqref{Projection-Free-eq1}, we have $\x_i^\ast(z-1)=\argmin_{\x\in\K}F_{z-1,i}(\x)$. Then, because of $\x_i(z)=\text{CG}(\K,L,F_{z-1,i}(\x),\x_i(z-1))$ and Lemma \ref{lem-CG}, for any $z=3,\dots,T/L$, we have
\begin{equation}
\label{Projection-Free-eq2}
\begin{split}
F_{z-1,i}(\x_i(z))-F_{z-1,i}(\x_i^\ast(z-1))\leq  \frac{8((z-3)L\alpha+2h) R^2}{L+2}.
\end{split}
\end{equation}
By combining \eqref{sub_eq5} with \eqref{Projection-Free-eq2}, for any $z=3,\dots,T/L$, we further have
\begin{equation}
\label{Projection-Free-eq3}
\|\x_i(z)-\x_i^\ast(z-1)\|_2\leq \sqrt{\frac{2\left(F_{z-1,i}(\x_i(z))-F_{z-1,i}(\x_i^\ast(z-1))\right)}{(z-3)L\alpha+2h}}\leq \frac{4R}{\sqrt{L+2}}.
\end{equation}
Then, to bound the second term in the right side of \eqref{Projection-Free-eq1}, we recall that $\theta$ and $L^\prime$ used in Algorithm \ref{Projection-Free} equal to $\theta$ and $L$ used in Algorithm \ref{ADOA}, respectively. Following the proof of Lemma \ref{lem-fastMix}, it is easy to verify that that Algorithm \ref{Projection-Free} also enjoys the error bound presented in Lemma \ref{lem-fastMix}, i.e.,
\[\left\|\z_i(z)-\bar{\z}(z)\right\|_2\leq 3L(G+\alpha R)\]
for any $i\in[n]$ and $z\in[T/L]$, though $L$ in Algorithm \ref{Projection-Free} is different from $L^\prime$. Therefore, we can directly utilize \eqref{sub_eq8} in the proof of Theorem \ref{sconvex_upper} to derive the following upper bound
\begin{equation}
\label{Projection-Free-eq4}
\|\x_i^\ast(z-1)-\bar{\x}(z)\|_2\leq \frac{3L(G+\alpha R)}{(z-3)L\alpha +2h}+\frac{4L(G+2\alpha R)}{(z-1)L\alpha+2h}.
\end{equation}
for any $z=3,\dots,T/L$. Moreover, it is not hard to verify that the third term in the right side of \eqref{Projection-Free-eq1} can be bounded by \eqref{sub_eq7} in the proof of Theorem \ref{sconvex_upper}.

Substituting \eqref{Projection-Free-eq3}, \eqref{Projection-Free-eq4}, and \eqref{sub_eq7} into \eqref{Projection-Free-eq1}, for any $z=3,\dots,T/L$, we have
\begin{equation*}
\begin{split}
&\|\x_i(z)-\bar{\x}(z+1)\|_2\\
\leq&\frac{4R}{\sqrt{L+2}}+\frac{3L(G+\alpha R)}{(z-3)L\alpha +2h}+\frac{4L(G+2\alpha R)}{(z-1)L\alpha+2h}+\frac{2L(G+2\alpha R)}{zL\alpha+2h}\\
\leq&\frac{4R}{\sqrt{L+2}}+\frac{3L(G+\alpha R)}{(z-3)L\alpha +2h}+\frac{6L(G+2\alpha R)}{(z-1)L\alpha+2h}.
\end{split}
\end{equation*}
Moreover, because of $\x_i(1)=\x_i(2)=\bar{\x}(1)=\ze$, it is easy to verify that
\begin{equation*}
\begin{split}
\|\x_i(z)-\bar{\x}(z+1)\|_2=\|\bar{\x}_i(z)-\bar{\x}(z+1)\|_2\overset{\eqref{sub_eq7}}{\leq}\frac{2L(G+2\alpha R)}{zL\alpha+2h}
\end{split}
\end{equation*}
for $z=1$, and
\begin{equation*}
\begin{split}
\|\x_i(z)-\bar{\x}(z+1)\|_2=\|\bar{\x}_i(z-1)-\bar{\x}(z+1)\|_2\overset{\eqref{sub_eq7}}{\leq}\frac{4L(G+2\alpha R)}{zL\alpha+2h}.
\end{split}
\end{equation*}
for $z=2$. Finally, following \eqref{for-projection-free-ref} in the proof of Theorem \ref{sconvex_upper}, it is easy to verify that
\begin{equation*}
\begin{split}
&\sum_{z=1}^{T/L}\sum_{t\in\mathcal{T}_z}\sum_{j=1}^nf_{t,j}(\x_i(z))-\sum_{z=1}^{T/L}\sum_{t\in\mathcal{T}_z}\sum_{j=1}^nf_{t,j}(\x)\\
\leq&nhR^2+\frac{12nGRT}{\sqrt{L+2}}+3nLG\left(\sum_{z=3}^{T/L}\frac{3L(G+\alpha R)}{(z-3)L\alpha +2h}+\sum_{z=1}^{T/L}\frac{6L(G+2\alpha R)}{(z-1)L\alpha+2h}\right).
\end{split}
\end{equation*}

\subsection{Proof of Corollaries \ref{corollary-convex_upper-projection-free} and \ref{corollary-sconvex_upper-projection-free}}
By substituting $\alpha=0$ into \eqref{unified-thm-pro}, we have
\begin{equation}
\label{cor3-eq1}
\begin{split}
R_{T,i}\leq&3nLG\left(\sum_{z=3}^{T/L}\frac{3LG}{2h}+\sum_{z=1}^{T/L}\frac{3LG}{h}\right)+nhR^2+\frac{12nGRT}{\sqrt{L+2}}\\
\leq&\frac{14nLG^2T}{h}+nhR^2+\frac{12nGRT}{\sqrt{L+2}}.
\end{split}
\end{equation}
Then, by substituting $h=\sqrt{14LT}G/R$ and $L=\sqrt{T}L^\prime$ into \eqref{cor3-eq1}, we have
\begin{equation}
\label{cor3-eq3}
\begin{split}
R_{T,i}
\leq&2\sqrt{14}nGR\sqrt{L^\prime}T^{3/4}+\frac{12nGRT^{3/4}}{\sqrt{L^\prime}}.
\end{split}
\end{equation}
If $L^\prime\leq \sqrt{T}$, by substituting  $h=\sqrt{14LT}G/R$ and $L=\sqrt{T}$ into \eqref{cor3-eq1}, we have
\begin{equation}
\label{cor3-eq2}
\begin{split}
R_{T,i}
\leq&(2\sqrt{14}+12)nGRT^{3/4}.
\end{split}
\end{equation}
From \eqref{cor3-eq3} and \eqref{cor3-eq2}, the proof of Corollary \ref{corollary-convex_upper} is completed.

To prove Corollary \ref{corollary-sconvex_upper-projection-free}, we first substitute $h=\alpha L$ into \eqref{unified-thm-pro} to derive the following regret bound
\begin{equation}
\label{cor4-eq1}
\begin{split}
R_{T,i}\leq& 3nLG\left(\sum_{z=3}^{T/L}\frac{3(G+\alpha R)}{(z-1)\alpha}+\sum_{z=1}^{T/L}\frac{6(G+2\alpha R)}{(z+1)\alpha}\right)+n\alpha LR^2+\frac{12nGRT}{\sqrt{L+2}}\\
\leq&\frac{3nLG(9G+15\alpha R)}{\alpha}\sum_{z=1}^{T/L}\frac{1}{z}+n\alpha LR^2+\frac{12nGRT}{\sqrt{L+2}}\\
\leq&\frac{3nLG(9G+15\alpha R)(1+\ln(T/L))}{\alpha}+n\alpha LR^2+\frac{12nGRT}{\sqrt{L+2}}.
\end{split}
\end{equation}
Then, by substituting $L=T^{2/3}(\ln T)^{-2/3}L^\prime$ into \eqref{cor4-eq1}, we have
\begin{equation}
\label{cor4-eq2}
\begin{split}
R_{T,i}
\leq&\frac{3nG(9G+15\alpha R)T^{2/3}L^\prime((\ln T)^{-2/3}+(\ln T)^{1/3})}{\alpha}\\
&+n\alpha T^{2/3}L^\prime R^2(\ln T)^{-2/3}+\frac{12nGRT^{2/3}(\ln T)^{1/3}}{\sqrt{L^\prime}}.
\end{split}
\end{equation}
Moreover, if $L^\prime\leq T^{2/3}(\ln T)^{-2/3}$, by substituting $L=T^{2/3}(\ln T)^{-2/3}$ into \eqref{cor4-eq1}, we have
\begin{equation}
\label{cor4-eq3}
\begin{split}
R_{T,i}
\leq&\frac{3nG(9G+15\alpha R)T^{2/3}((\ln T)^{-2/3}+(\ln T)^{1/3})}{\alpha}\\
&+n\alpha T^{2/3} R^2(\ln T)^{-2/3}+12nGRT^{2/3}(\ln T)^{1/3}.
\end{split}
\end{equation}
From \eqref{cor4-eq2} and \eqref{cor4-eq3}, the proof of Corollary \ref{corollary-sconvex_upper-projection-free} is completed.

\section{Conclusion}
This paper investigates D-OCO with convex and strongly convex functions, and aims to develop optimal and efficient algorithms. To this end, we first propose a novel D-OCO algorithm, namely AD-FTGL, which reduces the existing $O(n^{5/4}\rho^{-1/2}\sqrt{T})$ and $O(n^{3/2}\rho^{-1}\log T)$ regret bounds for convex and strongly convex functions to $\tilde{O}(n\rho^{-1/4}\sqrt{T})$ and $\tilde{O}(n\rho^{-1/2}\log T)$, respectively. Furthermore, we demonstrate its optimality for D-OCO by deriving $\Omega(n\rho^{-1/4}\sqrt{T})$ and $\Omega(n\rho^{-1/2} \log T)$ lower bounds for convex and strongly convex functions, respectively. Finally, to efficiently handle complex constraints, we propose a projection-free variant of AD-FTGL, which can respectively achieve ${O}(nT^{3/4})$ and ${O}(nT^{2/3}(\log T)^{1/3})$ regret bounds for convex and strongly convex functions with only $\tilde{O}(\rho^{-1/2}\sqrt{T})$ and $\tilde{O}(\rho^{-1/2}T^{1/3}(\log T)^{2/3})$ communication rounds. Although these regret bounds cannot match the aforementioned lower bounds, they are much tighter than those of existing projection-free algorithms for D-OCO. Moreover, we provide communication-dependent lower bounds to demonstrate that the number of communication rounds required by our projection-free algorithm is nearly optimal for achieving these regret bounds.


\vskip 0.2in
\bibliography{ref}

\begin{thebibliography}{50}
\providecommand{\natexlab}[1]{#1}
\providecommand{\url}[1]{\texttt{#1}}
\expandafter\ifx\csname urlstyle\endcsname\relax
  \providecommand{\doi}[1]{doi: #1}\else
  \providecommand{\doi}{doi: \begingroup \urlstyle{rm}\Url}\fi

\bibitem[Abernethy et~al.(2008)Abernethy, Bartlett, Rakhlin, and
  Tewari]{Abernethy08}
Jacob Abernethy, Peter~L. Bartlett, Alexander Rakhlin, and Ambuj Tewari.
\newblock Optimal strategies and minimax lower bounds for online convex games.
\newblock In \emph{Proceedings of the 21st Annual Conference on Learning
  Theory}, pages 415--423, 2008.

\bibitem[Awerbuch and Kleinberg(2004)]{Awerbuch04}
Baruch Awerbuch and Robert~D. Kleinberg.
\newblock Adaptive routing with end-to-end feedback: Distributed learning and
  geometric approaches.
\newblock In \emph{Proceedings of the 36th Annual ACM Symposium on Theory of
  Computing}, pages 45--53, 2004.

\bibitem[Awerbuch and Kleinberg(2008)]{Awerbuch2008}
Baruch Awerbuch and Robert~D. Kleinberg.
\newblock Online linear optimization and adaptive routing.
\newblock \emph{Journal of Computer and System Sciences}, 74\penalty0
  (1):\penalty0 97--114, 2008.

\bibitem[Boyd et~al.(2006)Boyd, Ghosh, Prabhakar, and Shah]{Gossip06}
Stephen Boyd, Arpita Ghosh, Balaji Prabhakar, and Devavrat Shah.
\newblock Randomized gossip algorithms.
\newblock \emph{IEEE Transactions on Information Theory}, 52\penalty0
  (6):\penalty0 2508--2530, 2006.

\bibitem[Cesa-Bianchi et~al.(2004)Cesa-Bianchi, Conconi, and Gentile]{O2B}
Nicol{\`o} Cesa-Bianchi, Alex Conconi, and Claudio Gentile.
\newblock On the generalization ability of on-line learning algorithms.
\newblock \emph{IEEE Transactions on Information Theory}, 50\penalty0
  (9):\penalty0 2050--2057, 2004.

\bibitem[Duchi et~al.(2011)Duchi, Agarwal, and Wainwright]{DADO2011}
John~C. Duchi, Alekh Agarwal, and Martin~J. Wainwright.
\newblock Dual averaging for distributed optimization: Convergence analysis and
  network scaling.
\newblock \emph{IEEE Transactions on Automatic Control}, 57\penalty0
  (3):\penalty0 592--606, 2011.

\bibitem[Dvinskikh and Gasnikov(2021)]{Dvinskikh-Arxiv-21}
Darina Dvinskikh and Alexander Gasnikov.
\newblock Decentralized and parallel primal and dual accelerated methods for
  stochastic convex programming problems.
\newblock \emph{Journal of Inverse and Ill-posed Problems}, 29\penalty0
  (3):\penalty0 385--405, 2021.

\bibitem[Frank and Wolfe(1956)]{FW-56}
Marguerite Frank and Philip Wolfe.
\newblock An algorithm for quadratic programming.
\newblock \emph{Naval Research Logistics Quarterly}, 3\penalty0
  (1--2):\penalty0 95--110, 1956.

\bibitem[Garber and Hazan(2016)]{Garber16}
Dan Garber and Elad Hazan.
\newblock A linearly convergent conditional gradient algorithm with
  applications to online and stochastic optimization.
\newblock \emph{SIAM Journal on Optimization}, 26\penalty0 (3):\penalty0
  1493--1528, 2016.

\bibitem[Garber and Kretzu(2020)]{Garber19}
Dan Garber and Ben Kretzu.
\newblock Improved regret bounds for projection-free bandit convex
  optimization.
\newblock In \emph{Proceedings of the 23rd International Conference on
  Artificial Intelligence and Statistics}, pages 2196--2206, 2020.

\bibitem[Garber and Kretzu(2021)]{Garber_SOFW}
Dan Garber and Ben Kretzu.
\newblock Revisiting projection-free online learning: the strongly convex case.
\newblock In \emph{Proceedings of the 24th International Conference on
  Artificial Intelligence and Statistics}, pages 3592--3600, 2021.

\bibitem[Garber and Kretzu(2022)]{Garber-COLT22}
Dan Garber and Ben Kretzu.
\newblock New projection-free algorithms for online convex optimization with
  adaptive regret guarantees.
\newblock In \emph{Proceedings of 35th Conference on Learning Theory}, pages
  326--2359, 2022.

\bibitem[Garber and Kretzu(2023)]{Garber-COLT23}
Dan Garber and Ben Kretzu.
\newblock Projection-free online exp-concave optimization.
\newblock In \emph{Proceedings of 36th Conference on Learning Theory}, pages
  1259--1284, 2023.

\bibitem[Gorbunov et~al.(2020)Gorbunov, Dvinskikh, and
  Gasnikov]{Gorbunov-Arxiv-2020}
Eduard Gorbunov, Darina Dvinskikh, and Alexander Gasnikov.
\newblock Optimal decentralized distributed algorithms for stochastic convex
  optimization.
\newblock \emph{arXiv:1911.07363}, 2020.

\bibitem[Hazan(2016)]{Hazan2016}
Elad Hazan.
\newblock Introduction to online convex optimization.
\newblock \emph{Foundations and Trends in Optimization}, 2\penalty0
  (3--4):\penalty0 157--325, 2016.

\bibitem[Hazan and Kale(2012)]{Hazan2012}
Elad Hazan and Satyen Kale.
\newblock Projection-free online learning.
\newblock In \emph{Proceedings of the 29th International Conference on Machine
  Learning}, pages 1843--1850, 2012.

\bibitem[Hazan and Kale(2014)]{Epoch-GD}
Elad Hazan and Satyen Kale.
\newblock Beyond the regret minimization barrier: Optimal algorithms for
  stochastic strongly-convex optimization.
\newblock \emph{Journal of Machine Learning Research}, 15\penalty0
  (71):\penalty0 2489--2512, 2014.

\bibitem[Hazan and Minasyan(2020)]{Hazan20}
Elad Hazan and Edgar Minasyan.
\newblock Faster projection-free online learning.
\newblock In \emph{Proceedings of the 33rd Annual Conference on Learning
  Theory}, pages 1877--1893, 2020.

\bibitem[Hazan et~al.(2007)Hazan, Agarwal, and Kale]{Hazan_2007}
Elad Hazan, Amit Agarwal, and Satyen Kale.
\newblock Logarithmic regret algorithms for online convex optimization.
\newblock \emph{Machine Learning}, 69\penalty0 (2):\penalty0 169--192, 2007.

\bibitem[Hosseini et~al.(2013)Hosseini, Chapman, and Mesbahi]{D-ODA}
Saghar Hosseini, Airlie Chapman, and Mehran Mesbahi.
\newblock Online distributed optimization via dual averaging.
\newblock In \emph{52nd IEEE Conference on Decision and Control}, pages
  1484--1489, 2013.

\bibitem[Jaggi(2013)]{Revist_FW}
Martin Jaggi.
\newblock Revisiting frank-wolfe: Projection-free sparse convex optimization.
\newblock In \emph{Proceedings of the 30th International Conference on Machine
  Learning}, pages 427--435, 2013.

\bibitem[Kovalev et~al.(2020)Kovalev, Salim, and Richtarik]{Kovalev-NIPS20}
Dmitry Kovalev, Adil Salim, and Peter Richtarik.
\newblock Optimal and practical algorithms for smooth and strongly convex
  decentralized optimization.
\newblock In \emph{Advances in Neural Information Processing Systems 33}, pages
  18342--18352, 2020.

\bibitem[Lei et~al.(2020)Lei, Yi, Hong, Chen, and Shi]{NIPS20-random-DOCO}
Jinlong Lei, Peng Yi, Yiguang Hong, Jie Chen, and Guodong Shi.
\newblock Online convex optimization over {Erd{\H o}s--R{\'e}nyi} random
  networks.
\newblock In \emph{Advances in Neural Information Processing Systems 33}, pages
  15591--15601, 2020.

\bibitem[Lesser et~al.(2003)Lesser, Ortiz, and Tambe]{DSN_book}
Victor Lesser, Charles~L. Ortiz, and Milind Tambe, editors.
\newblock \emph{Distributed Sensor Networks: A Multiagent Perspective}.
\newblock Springer New York, 2003.

\bibitem[Levy and Krause(2019)]{kevy_smooth}
Kfir~Y. Levy and Andreas Krause.
\newblock Projection free online learning over smooth sets.
\newblock In \emph{Proceedings of the 22nd International Conference on
  Artificial Intelligence and Statistics}, pages 1458--1466, 2019.

\bibitem[Li et~al.(2002)Li, Wong, Hu, and Sayeed]{Distrbuted02}
Dan Li, Kerry~D. Wong, Yu~H. Hu, and Akbar~M. Sayeed.
\newblock Detection, classification and tracking of targets in distributed
  sensor networks.
\newblock \emph{IEEE Signal Processing Magazine}, 19\penalty0 (2):\penalty0
  17--29, 2002.

\bibitem[Liu and Morse(2011)]{Acc_Gossip}
Ji~Liu and A.~Stephen Morse.
\newblock Accelerated linear iterations for distributed averaging.
\newblock \emph{Annual Reviews in Control}, 35\penalty0 (2):\penalty0 160--165,
  2011.

\bibitem[Lu and Sa(2021)]{Lu-ICML-21}
Yucheng Lu and Christopher~De Sa.
\newblock Optimal complexity in decentralized training.
\newblock In \emph{Proceedings of the 38th International Conference on Machine
  Learning}, pages 7111--7123, 2021.

\bibitem[Nesterov(2009)]{Dual-Averaging09}
Yurii Nesterov.
\newblock Primal-dual subgradient methods for convex problems.
\newblock \emph{Mathematical Programming}, 120\penalty0 (1):\penalty0 221--259,
  2009.

\bibitem[Scaman et~al.(2017)Scaman, Bach, Bubeck, Lee, and
  Massouli{{\'e}}]{Scaman-ICML17}
Kevin Scaman, Francis Bach, S{{\'e}}bastien Bubeck, Yin~Tat Lee, and Laurent
  Massouli{{\'e}}.
\newblock Optimal algorithms for smooth and strongly convex distributed
  optimization in networks.
\newblock In \emph{Proceedings of the 34th International Conference on Machine
  Learning}, pages 3027--3036, 2017.

\bibitem[Scaman et~al.(2018)Scaman, Bach, Bubeck, Lee, and
  Massouli{{\'e}}]{Scaman-NIPS18}
Kevin Scaman, Francis Bach, S{{\'e}}bastien Bubeck, Yin~Tat Lee, and Laurent
  Massouli{{\'e}}.
\newblock Optimal algorithms for non-smooth distributed optimization in
  networks.
\newblock In \emph{Advances in Neural Information Processing Systems 31}, pages
  2740--2749, 2018.

\bibitem[Scaman et~al.(2019)Scaman, Bach, Bubeck, Lee, and
  Massouli{{\'e}}]{Scaman-JMLR19}
Kevin Scaman, Francis Bach, S{{\'e}}bastien Bubeck, Yin~Tat Lee, and Laurent
  Massouli{{\'e}}.
\newblock Optimal convergence rates for convex distributed optimization in
  networks.
\newblock \emph{Journal of Machine Learning Research}, 20\penalty0
  (159):\penalty0 1--31, 2019.

\bibitem[Shalev-Shwartz(2011)]{Online:suvery}
Shai Shalev-Shwartz.
\newblock Online learning and online convex optimization.
\newblock \emph{Foundations and Trends in Machine Learning}, 4\penalty0
  (2):\penalty0 107--194, 2011.

\bibitem[Shalev-Shwartz and Singer(2007)]{Shai07}
Shai Shalev-Shwartz and Yoram Singer.
\newblock A primal-dual perspective of online learning algorithm.
\newblock \emph{Machine Learning}, 69\penalty0 (2--3):\penalty0 115--142, 2007.

\bibitem[Song et~al.(2023)Song, Shi, Pu, and Yan]{SongAccGossip23}
Zhuoqing Song, Lei Shi, Shi Pu, and Ming Yan.
\newblock Optimal gradient tracking for decentralized optimization.
\newblock \emph{Mathematical Programming}, pages 1--53, 2023.

\bibitem[Wan and Zhang(2021)]{SC_OFW}
Yuanyu Wan and Lijun Zhang.
\newblock Projection-free online learning over strongly convex sets.
\newblock In \emph{Proceedings of the 35th AAAI Conference on Artificial
  Intelligence}, pages 10076--10084, 2021.

\bibitem[Wan et~al.(2020)Wan, Tu, and Zhang]{Wan-ICML-2020}
Yuanyu Wan, Wei-Wei Tu, and Lijun Zhang.
\newblock Projection-free distributed online convex optimization with
  ${O}(\sqrt{T})$ communication complexity.
\newblock In \emph{Proceedings of the 37th International Conference on Machine
  Learning}, pages 9818--9828, 2020.

\bibitem[Wan et~al.(2021{\natexlab{a}})Wan, Wang, and Zhang]{Wan2021arXiv}
Yuanyu Wan, Guanghui Wang, and Lijun Zhang.
\newblock Projection-free distributed online learning with strongly convex
  losses.
\newblock \emph{arXiv:2103.11102v1}, 2021{\natexlab{a}}.

\bibitem[Wan et~al.(2021{\natexlab{b}})Wan, Xue, and Zhang]{DY_OFW}
Yuanyu Wan, Bo~Xue, and Lijun Zhang.
\newblock Projection-free online learning in dynamic environments.
\newblock In \emph{Proceedings of the 35th AAAI Conference on Artificial
  Intelligence}, pages 10067--10075, 2021{\natexlab{b}}.

\bibitem[Wan et~al.(2022)Wan, Wang, Tu, and Zhang]{Wan-22-JMLR}
Yuanyu Wan, Guanghui Wang, Wei-Wei Tu, and Lijun Zhang.
\newblock Projection-free distributed online learning with sublinear
  communication complexity.
\newblock \emph{Journal of Machine Learning Research}, 23\penalty0
  (172):\penalty0 1--53, 2022.

\bibitem[Wan et~al.(2023)Wan, Zhang, and Song]{COLT-23-wan}
Yuanyu Wan, Lijun Zhang, and Mingli Song.
\newblock Improved dynamic regret for online frank-wolfe.
\newblock In \emph{Proceedings of the 36th Annual Conference on Learning
  Theory}, pages 3304--3327, 2023.

\bibitem[Wan et~al.(2024)Wan, Wei, Song, and Zhang]{COLT-24-wan}
Yuanyu Wan, Tong Wei, Mingli Song, and Lijun Zhang.
\newblock Nearly optimal regret for decentralized online convex optimization.
\newblock In \emph{Proceedings of the 37th Annual Conference on Learning
  Theory}, pages 4862--4888, 2024.

\bibitem[Wang et~al.(2023)Wang, Wan, Zhang, and Zhang]{Wang-AAAI-2023}
Yibo Wang, Yuanyu Wan, Shimao Zhang, and Lijun Zhang.
\newblock Distributed projection-free online learning for smooth and convex
  losses.
\newblock In \emph{Proceedings of the 37th AAAI Conference on Artificial
  Intelligence}, pages 10226--10234, 2023.

\bibitem[Xiao(2009)]{NIPS2009_3882}
Lin Xiao.
\newblock Dual averaging method for regularized stochastic learning and online
  optimization.
\newblock In \emph{Advances in Neural Information Processing Systems 22}, pages
  2116--2124, 2009.

\bibitem[Xiao and Boyd(2004)]{Xiao-Gossip04}
Lin Xiao and Stephen Boyd.
\newblock Fast linear iterations for distributed averaging.
\newblock \emph{Systems and Control Letters}, 53\penalty0 (1):\penalty0 65--78,
  2004.

\bibitem[Yan et~al.(2013)Yan, Sundaram, Vishwanathan, and Qi]{DAOL_TKDE}
Feng Yan, Shreyas Sundaram, S.V.N. Vishwanathan, and Yuan Qi.
\newblock Distributed autonomous online learning: Regrets and intrinsic
  privacy-preserving properties.
\newblock \emph{IEEE Transactions on Knowledge and Data Engineering},
  25\penalty0 (11):\penalty0 2483--2493, 2013.

\bibitem[Ye and Chang(2023)]{YeAccGossip23}
Haishan Ye and Xiangyu Chang.
\newblock Optimal decentralized composite optimization for strongly convex
  functions.
\newblock \emph{arXiv:2312.15845}, 2023.

\bibitem[Ye et~al.(2023)Ye, Luo, Zhou, and Zhang]{Ye2020}
Haishan Ye, Luo Luo, Ziang Zhou, and Tong Zhang.
\newblock Multi-consensus decentralized accelerated gradient descent.
\newblock \emph{Journal of Machine Learning Research}, 24\penalty0
  (306):\penalty0 1--50, 2023.

\bibitem[Zhang et~al.(2017)Zhang, Zhao, Zhu, Hoi, and Zhang]{wenpeng17}
Wenpeng Zhang, Peilin Zhao, Wenwu Zhu, Steven C.~H. Hoi, and Tong Zhang.
\newblock Projection-free distributed online learning in networks.
\newblock In \emph{Proceedings of the 34th International Conference on Machine
  Learning}, pages 4054--4062, 2017.

\bibitem[Zinkevich(2003)]{Zinkevich2003}
Martin Zinkevich.
\newblock Online convex programming and generalized infinitesimal gradient
  ascent.
\newblock In \emph{Proceedings of the 20th International Conference on Machine
  Learning}, pages 928--936, 2003.

\end{thebibliography}

\newpage
\appendix

\section{Revisiting D-FTGL}
\label{appendix_last}
\citet{Wan-22-JMLR} originally develop a projection-free version of D-FTGL for $\alpha$-strongly convex functions, which also adopts the blocking update mechanism to update the decision of each learner. In this section, we first discuss how to simplify the projection-free version into D-FTGL, and then provide a refined analysis of its regret.
\subsection{The Algorithm}
Following the notations in our Algorithm \ref{ADOA}, at the end of each block $z=1,\dots,T/L$, each learner $i$ of their algorithm updates as
\begin{equation}
 \label{DFTAL-projection-free}
\begin{split}
&\z_{i}(z+1)=\sum_{j\in N_i}P_{ij}\z_j(z)+\d_i(z)\\
&\x_i(z+1) = \text{CG}(\K,K,F_{z,i}(\x),\x_i(z))
\end{split}
\end{equation}
where $F_{z,i}(\x)=\left\langle\z_{i}(z),\x\right\rangle+\frac{(z-1)L\alpha}{2}\|\x\|_2^2+h\|\x\|_2^2$, and $\x_i(z+1)$ is computed by utilizing CG \citep{FW-56,Revist_FW} with the initialization $\x_i(z)$ and $K$ iterations to minimize the function $F_{z,i}(\x)$ over the decision set $\K$.

According to Theorem 2 of \citet{Wan-22-JMLR}, under Assumptions \ref{assum5}, \ref{assum4}, \ref{assum1}, and \ref{scvx-assum}, their algorithm can achieve the following regret bound
\begin{equation}
 \label{regret-DFTAL-projection-free}
 R_{T,i}\leq \frac{12nGRT}{\sqrt{K+2}}+\sum_{z=2}^{T/L}\frac{3nGL^2(G+\alpha R)\sqrt{n}}{((z-2)\alpha L+2h)(1-\sigma_2(P))}+\sum_{z=1}^{T/L}\frac{4nL^2(G+2\alpha R)^2}{z\alpha L+2h}+4nhR^2.
\end{equation}
By substituting $K=L=\sqrt{T}$, $\alpha=0$, and $h=n^{1/4}T^{3/4}G/(\sqrt{\rho}R)$ into \eqref{regret-DFTAL-projection-free}, they derive a regret bound of ${O}(n^{5/4}\rho^{-1/2}T^{3/4})$ for convex functions. Moreover, by substituting $K=L=T^{2/3}(\ln T)^{-2/3}$ and $h=\alpha L$ into \eqref{regret-DFTAL-projection-free}, they derive a regret bound of ${O}(n^{3/2}\rho^{-1}T^{2/3}(\log T)^{1/3})$ for strongly convex functions.

However, these choices of $K$ and $L$ are utilized for achieving the projection-free property, i.e., only one linear optimization step is utilized per round on average. Actually, it is easy to derive the ${O}(n^{5/4}\rho^{-1/2}\sqrt{T})$ regret bound for convex functions by substituting $K=\infty$, $L=1$, and $h=n^{1/4}\sqrt{T}G/(\sqrt{\rho}R)$, and  the ${O}(n^{3/2}\rho^{-1}\log T)$ regret bound for strongly convex functions by substituting $K=\infty$, $L=1$, and $h=\alpha$ into \eqref{regret-DFTAL-projection-free}. With the new choice of $K$ and $L$, the algorithm of \citet{Wan-22-JMLR} reduces to performing the following update
\begin{equation}
 \label{DFTAL-Lazy}
\begin{split}
&\z_{i}(t+1)=\sum_{j\in N_i}P_{ij}\z_j(t)+(\nabla f_{t,i}(\x_i(t))-\alpha\x_i(t))\\
&\x_i(t+1) = \argmin_{\x\in\K}\left\langle\z_{i}(t),\x\right\rangle+\frac{(t-1)\alpha}{2}\|\x\|_2^2+h\|\x\|_2^2
\end{split}
\end{equation}
for each learner $i$ at round $t$. Additionally, it is worth noting that the main reason for computing $\x_i(z+1)$ in \eqref{DFTAL-projection-free} based on $\z_i(z)$ is to allocate $K$ iterations of CG to every round in block $z$, rather than only the last round in the block. In contrast, here computing $\x_i(t+1)$ based on $\z_{i}(t)$ provides no benefit because of $L=1$. Thus, \eqref{DFTAL-Lazy} can be further simplified to D-FTGL, as given in \eqref{DFTAL}. For the sake of completeness, we summarize the detailed procedure in Algorithm \ref{D-FTGL-code}.
\begin{algorithm}[t]
\caption{D-FTGL}
\label{D-FTGL-code}
\begin{algorithmic}[1]
\STATE \textbf{Input:} $\alpha$, $h$ 
\STATE \textbf{Initialization:} set $\mathbf{x}_i(1)=\z_i(1)=\ze, \forall i\in [n]$
\FOR{$t=1,\dots,T$}
\FOR{each local learner $i\in [n]$}
\STATE Play $\x_{i}(t)$ and query $\nabla f_{t,i}(\x_{i}(t))$
\STATE Set $\z_{i}(t+1)=\sum_{j\in N_i}P_{ij}\z_j(t)+(\nabla f_{t,i}(\x_i(t))-\alpha\x_i(t))$
\STATE Compute $\x_i(z+1)=\argmin_{\x\in\K}\left\langle\z_{i}(t+1),\x\right\rangle+\frac{t\alpha}{2}\|\x\|_2^2+h\|\x\|_2^2$
\ENDFOR
\ENDFOR
\end{algorithmic}
\end{algorithm}

\subsection{Theoretical Guarantees}
Although D-FTGL is slightly different from \eqref{DFTAL-Lazy}, it is easy to verify that D-FTGL can also achieve the aforementioned regret bounds for convex and strongly convex functions. To be precise, we first establish the following guarantee for D-FTGL.
\begin{thm}
\label{general_upper_D-FTGL}
Let $C$ denote an upper bound of the approximate error of the standard gossip step in Algorithm \ref{D-FTGL-code}, i.e., $\|\z_i(t)-\bar{\z}(t)\|_2\leq C$ for any $t\in[T]$ and $i\in[n]$, where $\bar{\z}(t)=\frac{1}{n}\sum_{i=1}^n\z_i(t)$. Under Assumptions \ref{assum5}, \ref{assum4}, \ref{assum1}, and \ref{scvx-assum}, for any $i\in [n]$, Algorithm \ref{D-FTGL-code} ensures
\begin{equation}
\label{unified-thm-D-FTGL}
R_{T,i}\leq 3nG\left(\sum_{t=2}^{T}\frac{C}{(t-1)\alpha+2h}+\sum_{t=1}^{T}\frac{2(G+2\alpha R)}{t\alpha+2h}\right)+nhR^2.
\end{equation}
\end{thm}
Note that Lemma 3 of \citet{Wan-22-JMLR} has provided an error bound of $C=\sqrt{n}(G+\alpha R)\rho^{-1}$ for the standard gossip step. By substituting this bound into \eqref{unified-thm-D-FTGL}, we can set $h=n^{1/4}\sqrt{T}G/(\sqrt{\rho}R)$ and $\alpha=0$ to achieve the ${O}(n^{5/4}\rho^{-1/2}\sqrt{T})$ regret bound for convex functions, and simply set $h=0$ to achieve the ${O}(n^{3/2}\rho^{-1}\log T)$ regret bound for strongly convex functions.

More importantly, we want to emphasize that the power of D-FTGL extends beyond recovering existing results. In the following, we provide a novel and improved error bound for the standard gossip step in D-FTGL, and further derive tighter regret bounds.
\begin{lem}
\label{impro-lem-gossip}
For any $i\in [n]$, let $\nabla_i(1),\dots,\nabla_i(T)\in\mathbb{R}^d$ be a sequence of vectors. Let $\z_i(1)=\ze$, $\z_{i}(t+1)=\sum_{j\in N_i}P_{ij}\z_{j}(t)+\nabla_{i}(t)$, and $\bar{\z}(t)=\frac{1}{n}\sum_{i=1}^n\z_i(t)$ for $t\in[T]$, where $P$ satisfies Assumption \ref{assum5}. For any $i\in [n]$ and $t\in[T]$, assuming $\|\nabla_i(t)\|_2\leq \xi$ where $\xi>0$ is a constant, we have \[\|\z_i(t)-\bar{\z}(t)\|_2\leq2\xi\left(\frac{1+\ln \sqrt{n}}{1-\sigma_2(P)}+1\right).\]
\end{lem}
This lemma is inspired by an existing error bound of $O(\rho^{-1}\log (nT))$ for the standard gossip step in decentralized offline optimization \citep{DADO2011}. However, possibly because of the additional dependence on $\log T$, it is overlooked in D-OCO, where $T$ could be very large and even the $\log T$ factor is unacceptable. In contrast, Lemma \ref{impro-lem-gossip} provides an $O(\rho^{-1}\log n)$ error bound without the $\log T$ factor. Moreover, compared with the $O(\sqrt{n}\rho^{-1})$ error bound in previous studies on D-OCO \citep{D-ODA,Wan-22-JMLR}, our error bound has a much tighter dependence on $n$. Then, by substituting $C=O(\rho^{-1}\log n)$, $h=G\sqrt{T\ln n}/(\sqrt{\rho}R)$ and $\alpha=0$ into \eqref{unified-thm-D-FTGL}, we can achieve an ${O}(n\rho^{-1/2}\sqrt{T\log n})$ regret bound for convex functions. Recall that D-FTGL with $\alpha=0$ reduces to D-FTRL \citep{D-ODA}. Here, the improved error bound allows us to tune the parameter $h$ better, and thus improve the existing ${O}(n^{5/4}\rho^{-1/2}\sqrt{T})$ regret bound. If functions are strongly convex, we can substitute $C=O(\rho^{-1}\log n)$ and $h=0$ into \eqref{unified-thm-D-FTGL} to derive an ${O}(n\rho^{-1}(\log n)\log T)$ regret bound, which is also tighter than the existing bound.
\begin{remark}
\emph{
It is worth noting that besides D-FTGL, other existing D-OCO algorithms with a similar use of the standard gossip step can benefit from our improved error bound in Lemma \ref{impro-lem-gossip}. For example, it is easy to verify that the ${O}(n^{5/4}\rho^{-1/2}T^{3/4})$ and ${O}(n^{3/2}\rho^{-1}T^{2/3}(\log T)^{1/3})$ regret bounds of the projection-free algorithm in \citet{Wan-22-JMLR} can be reduced to ${O}(n\rho^{-1/2}T^{3/4}\sqrt{\log n})$ and ${O}(n\rho^{-1}T^{2/3}(\log T)^{1/3}\log n)$, respectively.
}
\end{remark}
\subsection{Analysis}
In the following, we provide the detailed proofs of Theorem \ref{general_upper_D-FTGL} and Lemma \ref{impro-lem-gossip}.
\subsubsection{Proof of Theorem \ref{general_upper_D-FTGL}}
We start this proof by defining a virtual decision for any $t\in[T+1]$ as
\begin{equation}
\label{virtual_decision_scvx-D-FTGL}
\bar{\x}(t)=\argmin_{\x\in\K}\left\langle\x,\bar{\z}(t)\right\rangle+\frac{(t-1)\alpha}{2}\|\x\|_2^2+h\|\x\|_2^2
\end{equation}
For brevity, let $\d_i(t)=\nabla f_{t,i}(\x_i(t))-\alpha\x_i(t)$ and $\bar{\d}(t)=\frac{1}{n}\sum_{i=1}^n\d_i(t)$. In the following, we will bound the regret of any learner $i$ by analyzing the regret of $\bar{\x}(2),\dots,\bar{\x}(T+1)$ on a sequence of loss functions defined by $\bar{\d}(1),\dots,\bar{\d}(T)$ and the distance $\|\x_i(t)-\bar{\x}(t+1)\|_2$ for any $t\in[T]$. To this end, we notice that
\begin{equation}
\label{equal4Gossip}
\begin{split}
\bar{\z}(t+1)=&\frac{1}{n}\sum_{i=1}^n\z_i(t+1)=\frac{1}{n}\sum_{i=1}^n\left(\sum_{j\in N_i}P_{ij}\z_j(t)+\d_i(t)\right)\\
=&\frac{1}{n}\sum_{i=1}^n\sum_{j=1}^nP_{ij}\z_j(t)+\bar{\d}(t)=\frac{1}{n}\sum_{j=1}^n\sum_{i=1}^nP_{ij}\z_j(t)+\bar{\d}(t)\\
=&\bar{\z}(t)+\bar{\d}(t)=\sum_{i=1}^{t}\bar{\d}(i)
\end{split}
\end{equation}
where the third and last equalities are due to Assumption \ref{assum5}.

Then, let $\ell_t(\x)=\langle \x,\bar{\d}(t)\rangle+\frac{\alpha}{2}\|\x\|_2^2$ for any $t\in[T]$. By combining Lemma \ref{lem-ftl} with \eqref{virtual_decision_scvx-D-FTGL} and \eqref{equal4Gossip}, for any $\x\in\K$, it is easy to verify that
\begin{equation}
\label{sub_eq1-D-FTGL}
\begin{split}
\sum_{t=1}^{T}\ell_t(\bar{\x}(t+1))-\sum_{t=1}^{T}\ell_t(\x)\leq h\left(\|\x\|_2^2-\|\bar{\x}(2)\|_2^2\right)\leq hR^2
\end{split}
\end{equation}
where the last inequality is due to Assumption \ref{assum1} and $\|\bar{\x}(2)\|_2^2\geq0$. 

We also notice that for any $t\in[T]$, Algorithm \ref{D-FTGL-code} ensures
\begin{equation}
\label{pract_decision_scvx-D-FTGL}
\x_i(t)=\argmin_{\x\in\K}\left\langle\x, \z_i(t)\right\rangle+\frac{(t-1)\alpha}{2}\|\x\|_2^2+h\|\x\|_2^2.
\end{equation}
By combining Lemma \ref{lem-stab} with \eqref{virtual_decision_scvx-D-FTGL} and \eqref{pract_decision_scvx-D-FTGL}, for any $t\in[T]$, we have
\begin{equation}
\label{sub_eq4-D-FTGL}
\begin{split}
\|\x_i(t)-\bar{\x}(t)\|_2\leq&\frac{\left\|\z_i(t)-\bar{\z}(t)\right\|_2}{(t-1)\alpha+2h}\leq\frac{C}{(t-1)\alpha+2h}
\end{split}
\end{equation}
where the last inequality is due to the definition of $C$.

To bound $\|\x_i(t)-\bar{\x}(t+1)\|_2$, we still need to analyze the term $\|\bar{\x}(t)-\bar{\x}(t+1)\|_2$ for any $t\in[T]$. Let $F_t(\x)=\sum_{\tau=1}^{t}\ell_{\tau}(\x)+h\|\x\|_2^2$ for any $t\in[T]$. It is easy to verify that $F_t(\x)$
is $(t\alpha+2h)$-strongly convex over $\K$, and $\bar{\x}(t+1)=\argmin_{\x\in\K} F_t(\x)$. Then, we have
\begin{equation*}
\begin{split}
\|\bar{\x}(t)-\bar{\x}(t+1)\|_2^2\overset{\eqref{sub_eq5}}{\leq}&\frac{2}{t\alpha+2h}(F_t(\bar{\x}(t))-F_t(\bar{\x}(t+1)))\\
=&\frac{2}{t\alpha+2h}\left(F_{t-1}(\bar{\x}(t))-F_{t-1}(\bar{\x}(t+1))+\ell_t(\bar{\x}(t))-\ell_t(\bar{\x}(t+1))\right)\\
\overset{\eqref{virtual_decision_scvx-D-FTGL}}{\leq}&\frac{2(\ell_t(\bar{\x}(t))-\ell_t(\bar{\x}(t+1)))}{t\alpha+2h}{\leq}\frac{2(G+2\alpha R)\|\bar{\x}(t)-\bar{\x}(t+1)\|_2}{t\alpha+2h}
\end{split}
\end{equation*}
where the last inequality is due to
\begin{equation*}
\begin{split}
|\ell_t(\x)-\ell_t(\y)|\leq&\left|\langle \nabla \ell_t(\x),\x-\y\rangle\right|\leq\|\nabla \ell_t(\x)\|_2\|\x-\y\|_2\\
=&\|\bar{\d}(t)+\alpha \x\|_2\|\x-\y\|_2\leq\left(\left\|\frac{1}{n}\sum_{i=1}^n\d_i(t)\right\|_2+\|\alpha \x\|_2\right)\|\x-\y\|_2\\
{\leq}&(G+2\alpha R)\|\x-\y\|_2
\end{split}
\end{equation*}
for any $\x,\y\in\K$. Moreover, the above inequality can be simplified to
\begin{equation}
\begin{split}
\label{sub_eq7-D-FTGL}
\|\bar{\x}(t)-\bar{\x}(t+1)\|_2\leq\frac{2(G+2\alpha R)}{t\alpha+2h}.
\end{split}
\end{equation}
By combining (\ref{sub_eq4-D-FTGL}) and (\ref{sub_eq7-D-FTGL}), for any $t\in[T]$, we have
\begin{equation}
\label{sub_eq8-D-FTGL}
\begin{split}
\|\x_i(t)-\bar{\x}(t+1)\|_2\leq&\|\x_i(t)-\bar{\x}(t)\|_2+ \|\bar{\x}(t)-\bar{\x}(t+1)\|_2\\
\leq&\frac{C}{(t-1)\alpha+2h}+\frac{2(G+2\alpha R)}{t\alpha+2h}.
\end{split}
\end{equation}
It is also worth noting that $\x_i(1)=\bar{\x}(1)=\ze$, which ensures
\begin{equation}
\label{sub_eq8-D-FTGL-v2}
\begin{split}
\|\x_i(1)-\bar{\x}(2)\|_2=\|\bar{\x}(1)-\bar{\x}(2)\|_2\leq&\frac{2(G+2\alpha R)}{t\alpha+2h}.
\end{split}
\end{equation}
Now, we are ready to derive the regret bound of any learner $i$. For brevity, let $\epsilon_t$ denote the upper bound of $\|\x_i(t)-\bar{\x}(t+1)\|_2$ derived in \eqref{sub_eq8-D-FTGL} and \eqref{sub_eq8-D-FTGL-v2}. Similar to \eqref{for_ref-D-FTGL}, for any $t\in[T]$, $j\in [n]$, and $\x\in\K$, it is easy to verify that
\begin{equation*}
\begin{split}
f_{t,j}(\x_i(t))-f_{t,j}(\x)
\leq\langle\nabla f_{t,j}(\x_j(t)),\bar{\x}(t+1)-\x\rangle-\frac{\alpha}{2}\|\x_j(t)-\x\|_2^2+3G\epsilon_t.
\end{split}
\end{equation*}
By combining the above inequality with \eqref{for_ref-D-FTGL2}, for any $t\in[T]$, $j\in [n]$, and $\x\in\K$, we have
\begin{equation*}
\begin{split}
&f_{t,j}(\x_i(t))-f_{t,j}(\x)\\
\leq&\langle \nabla f_{t,j}(\x_j(t)),\bar{\x}(t+1)-\x\rangle-\frac{\alpha}{2}\left(2\langle \x_j(t),\bar{\x}(t+1)-\x\rangle+\|\x\|_2^2-\|\bar{\x}(t+1)\|_2^2\right)+3G\epsilon_t\\
=&\langle\nabla f_{t,j}(\x_j(t))-\alpha \x_j(t),\bar{\x}(t+1)-\x\rangle+\frac{\alpha}{2}\left(\|\bar{\x}(t+1)\|_2^2-\|\x\|_2^2\right)+3G\epsilon_t.
\end{split}
\end{equation*}
Finally, from the above inequality and the definition of $\epsilon_t$, it is not hard to verify that
\begin{equation*}
\begin{split}
&\sum_{t=1}^T\sum_{j=1}^nf_{t,j}(\x_i(t))-\sum_{t=1}^T\sum_{j=1}^nf_{t,j}(\x)\\
\leq&\sum_{t=1}^T\sum_{j=1}^n\left(\langle\nabla f_{t,j}(\x_j(t))-\alpha \x_j(t),\bar{\x}(t+1)-\x\rangle+\frac{\alpha}{2}\left(\|\bar{\x}(t+1)\|_2^2-\|\x\|_2^2\right)+3G\epsilon_t\right)\\
=&n\sum_{t=1}^{T}\left(\langle\bar{\d}(t),\bar{\x}(t+1)-\x\rangle+\frac{\alpha}{2}\left(\|\bar{\x}(t+1)\|_2^2-\|\x\|_2^2\right)\right)+3nG\sum_{t=1}^{T}\epsilon_t\\
\overset{\eqref{sub_eq1-D-FTGL}}{\leq}&nhR^2+3nG\left(\sum_{t=2}^{T}\frac{C}{(t-1)\alpha+2h}+\sum_{t=1}^{T}\frac{2(G+2\alpha R)}{t\alpha+2h}\right).
\end{split}
\end{equation*}

\subsubsection{Proof of Lemma \ref{impro-lem-gossip}}
This lemma is derived by refining the existing analysis for the standard gossip step \citep{DADO2011,D-ODA,wenpeng17,Wan-22-JMLR}. Let $P^{s}$ denote the $s$-th power of $P$ and $P^{s}_{ij}$ denote the $j$-th entry of the $i$-th row in $P^{s}$ for any $s\geq0$. Note that $P^0$ denotes the identity matrix $I_n$.
For $t=1$, it is easy to verify that
\begin{equation}
\label{eq0_zbound}
\|\z_i(t)-\bar{\z}(t)\|_2=0.
\end{equation}
To analyze the case with $T\geq t\geq2$, we introduce two intermediate results from \citet{wenpeng17} and \citet{DADO2011}. 

First, as shown in the proof of Lemma 6 at \citet{wenpeng17}, under Assumption \ref{assum5}, for any $T\geq t\geq2$, we have
\begin{equation}
\label{eq1_zbound}
\begin{split}
\|\z_i(t)-\bar{\z}(t)\|_2
\leq\sum_{\tau=1}^{t-1}\sum_{j=1}^n\left|P_{ij}^{t-1-\tau}-\frac{1}{n}\right|\|\nabla_j(\tau)\|_2.
\end{split}
\end{equation}
Second, as shown in Appendix B of \citet{DADO2011}, when $P$ is a doubly stochastic matrix, for any positive integer $s$ and any $\x$ in the $n$-dimensional probability simplex, it holds that
\begin{equation}
\label{eq3_zbound}
\|P^s\x-\mathbf{1}/n\|_1\leq \sigma_2(P)^s\sqrt{n}
\end{equation}
where $\mathbf{1}$ is the all-ones vector in $\mathbb{R}^n$.

Let $\mathbf{e}_i$ denote the $i$-th canonical basis vector in $\mathbb{R}^n$. By substituting $\x=\mathbf{e}_i$ into (\ref{eq3_zbound}), we have
\begin{equation}
\label{eq4_zbound}
\|P^s\mathbf{e}_i-\mathbf{1}/n\|_1\leq \sigma_2(P)^s\sqrt{n}
\end{equation}
for any positive integer $s$. If $s=0$, we also have
\begin{equation}
\label{eq5_zbound}
\|P^0\mathbf{e}_i-\mathbf{1}/n\|_1=\frac{2(n-1)}{n}\leq\sqrt{n}=\sigma_2(P)^0\sqrt{n}
\end{equation}
where the inequality is due to $n\geq1$.

Moreover, by combining (\ref{eq1_zbound}) with $\|\nabla_i(t)\|_2\leq \xi$, for any $T\geq t\geq2$, we have
\begin{equation}
\label{eq6_zbound}
\begin{split}
\|\z_i(t)-\bar{\z}(t)\|_2
\leq&\xi\sum_{\tau=1}^{t-1}\sum_{j=1}^n\left|P_{ij}^{t-1-\tau}-\frac{1}{n}\right|\\
=&\xi\sum_{\tau=1}^{t-1}\sum_{j=1}^n\left|P_{ji}^{t-1-\tau}-\frac{1}{n}\right|\\
=&\xi\sum_{\tau=1}^{t-1}\left\|P^{t-1-\tau}\mathbf{e}_i-\frac{\mathbf{1}}{n}\right\|_1
\end{split}
\end{equation}
where the first equality is due to the symmetry of $P$. To further bound the right side of \eqref{eq6_zbound}, previous studies have provided two different choices. 

First, as in \citet{DADO2011}, one can divide $\tau\in[1,t-1]$ into two parts by
\begin{equation}
\label{eq7_zbound}
\tau^\prime=t-1-\left\lceil\frac{\ln (T\sqrt{n})}{\ln \sigma_2(P)^{-1}}\right\rceil.
\end{equation}
For any $\tau\in[1,\tau^\prime]$, it is not hard to verify that
\begin{equation}
\label{eq8_zbound}
\left\|P^{t-1-\tau}\mathbf{e}_i-\frac{\mathbf{1}}{n}\right\|_1\overset{\eqref{eq4_zbound}}{\leq} \sigma_2(P)^{t-1-\tau}\sqrt{n}\overset{\eqref{eq7_zbound}}{\leq}\sigma_2(P)^{\frac{\ln (T\sqrt{n})}{\ln \sigma_2(P)^{-1}}}\sqrt{n}=\frac{1}{T}.
\end{equation}
Then, by combining \eqref{eq6_zbound} and \eqref{eq8_zbound}, for any $T\geq t\geq2$, it is easy to verify that
\begin{equation}
\label{eq9_zbound}
\begin{split}
\|\z_i(t)-\bar{\z}(t)\|_2
\leq&\frac{\xi\tau^\prime}{T}+\xi\sum_{\tau=\tau^\prime+1}^{t-1}\left\|P^{t-1-\tau}\mathbf{e}_i-\frac{\mathbf{1}}{n}\right\|_1\\
\leq&\xi+2\xi\left\lceil\frac{\ln (T\sqrt{n})}{\ln \sigma_2(P)^{-1}}\right\rceil\\
\leq &\xi+2\xi\left(\frac{\ln (T\sqrt{n})}{1-\sigma_2(P)}+1\right)
\end{split}
\end{equation}
where the second inequality is due to $\tau^\prime\leq T$ and $\left\|P^{t-1-\tau}\mathbf{e}_i-\frac{\mathbf{1}}{n}\right\|_1\leq2$ for any $\tau\leq t-1$, and the last inequality is due to $\ln x^{-1}\geq 1-x$ for any $x>0$.

The second choice is to simply combine \eqref{eq6_zbound} with (\ref{eq4_zbound}) and (\ref{eq5_zbound}) as in \citet{D-ODA}, \citet{wenpeng17}, and \citet{Wan-22-JMLR}, which provides the following upper bound
\begin{equation}
\label{eq10_zbound}
\begin{split}
\|\z_i(t)-\bar{\z}(t)\|_2
\leq\xi\sum_{\tau=1}^{t-1}\sigma_2(P)^{t-1-\tau}\sqrt{n}
=\frac{(1-\sigma_2(P)^{t-1})\xi\sqrt{n}}{1-\sigma_2(P)}\leq\frac{\xi\sqrt{n}}{1-\sigma_2(P)}
\end{split}
\end{equation}
for any $T\geq t\geq2$.

However, both bounds in \eqref{eq9_zbound} and \eqref{eq10_zbound} are unsatisfactory, since the former has a factor of $\ln T$ and the latter has a factor of $\sqrt{n}$. To address these issues, we incorporate the above two ideas. Specifically, we redefine $\tau^\prime$ as
\begin{equation}
\label{eq11_zbound}
\tau^\prime=t-1-\left\lceil\frac{\ln(\sqrt{n})}{\ln \sigma_2(P)^{-1}}\right\rceil.
\end{equation}
Then, from \eqref{eq6_zbound}, for any $T\geq t\geq2$, we have
\begin{equation}
\label{eq12_zbound}
\begin{split}
\|\z_i(t)-\bar{\z}(t)\|_2
\leq&\xi\sum_{\tau=1}^{\tau^\prime}\left\|P^{t-1-\tau}\mathbf{e}_i-\frac{\mathbf{1}}{n}\right\|_1+\xi\sum_{\tau=\tau^\prime+1}^{t-1}\left\|P^{t-1-\tau}\mathbf{e}_i-\frac{\mathbf{1}}{n}\right\|_1\\
\overset{\eqref{eq4_zbound}}{\leq}&\xi\sum_{\tau=1}^{\tau^\prime}\sigma_2(P)^{t-1-\tau}\sqrt{n}+\xi\sum_{\tau=\tau^\prime+1}^{t-1}\left\|P^{t-1-\tau}\mathbf{e}_i-\frac{\mathbf{1}}{n}\right\|_1\\
\leq&\xi\sum_{\tau=1}^{\tau^\prime}\sigma_2(P)^{\tau^\prime-\tau}+\xi\sum_{\tau=\tau^\prime+1}^{t-1}\left\|P^{t-1-\tau}\mathbf{e}_i-\frac{\mathbf{1}}{n}\right\|_1\\
=&\frac{(1-\sigma_2(P)^{\tau^\prime})\xi}{1-\sigma_2(P)}+\xi\sum_{\tau=\tau^\prime+1}^{t-1}\left\|P^{t-1-\tau}\mathbf{e}_i-\frac{\mathbf{1}}{n}\right\|_1\\
\leq&\frac{\xi}{1-\sigma_2(P)}+2\xi\left(\frac{\ln \sqrt{n}}{1-\sigma_2(P)}+1\right)
\end{split}
\end{equation}
where the third inequality is due to $\sigma_2(P)^{t-1-\tau^\prime}\leq 1/\sqrt{n}$ for $\tau^\prime$ in \eqref{eq11_zbound}, and the last inequality is due to $\left\|P^{t-1-\tau}\mathbf{e}_i-\frac{\mathbf{1}}{n}\right\|_1\leq2$ for any $\tau\leq t-1$ and \eqref{eq11_zbound}. Finally, this proof can be completed by combining \eqref{eq0_zbound} and \eqref{eq12_zbound}.

\end{document}